\documentclass[format=acmsmall, screen=true, nonacm]{acmart}

\AtBeginDocument{%
  }

\setcopyright{acmcopyright}
\copyrightyear{XXXX}
\acmYear{XXXX}
\acmDOI{XXXXXXX.XXXXXXX}

\acmJournal{TIST}
\acmVolume{0}
\acmNumber{0}
\acmArticle{0}
\acmMonth{0}

\usepackage{amsmath}
\usepackage{dirtytalk}
\usepackage{booktabs}
\usepackage{multirow}
\usepackage{subfig}
\usepackage{graphicx}
\usepackage{textcomp}
\usepackage{longtable}

\newcommand{\citeb}[1]{\textcolor{blue}{\cite{#1}}}

\begin{document}

%%
%% The "title" command has an optional parameter,
%% allowing the author to define a "short title" to be used in page headers.
% \title{Innovations in Neural Data-to-text Generation: A Survey}

\title{Innovations in Neural Data-to-text Generation: A Survey}

%%
%% The "author" command and its associated commands are used to define
%% the authors and their affiliations.
%% Of note is the shared affiliation of the first two authors, and the
%% "authornote" and "authornotemark" commands
%% used to denote shared contribution to the research.
\author{Mandar Sharma}
\orcid{0000-0002-7012-9323}
\affiliation{%
  \institution{Virginia Tech}
  \country{USA}}
\email{mandarsharma@vt.edu}

\author{Ajay Kumar Gogineni}
\affiliation{%
  \institution{Virginia Tech}
  \country{USA}}

\author{Naren Ramakrishnan}
\affiliation{%
  \institution{Virginia Tech}
  \country{USA}}

%%
%% By default, the full list of authors will be used in the page
%% headers. Often, this list is too long, and will overlap
%% other information printed in the page headers. This command allows
%% the author to define a more concise list
%% of authors' names for this purpose.
\renewcommand{\shortauthors}{Sharma et al.}

%%
%% The abstract is a short summary of the work to be presented in the
%% article.
\begin{abstract}
The neural boom that has sparked natural language processing (NLP) research throughout the last decade has similarly led to significant innovations in data-to-text generation (D2T). This survey offers a consolidated view into the neural D2T paradigm with a structured examination of the approaches, benchmark datasets, and evaluation protocols. This survey draws boundaries separating D2T from the rest of the natural language generation (NLG) landscape, encompassing an up-to-date synthesis of the literature, and highlighting the stages of technological adoption from within and outside the greater NLG umbrella. With this holistic view, we highlight promising avenues for D2T research that not only focus on the design of linguistically capable systems but also systems that exhibit fairness and accountability.
\end{abstract}

%%
%% The code below is generated by the tool at http://dl.acm.org/ccs.cfm.
%% Please copy and paste the code instead of the example below.
%%


%%
%% Keywords. The author(s) should pick words that accurately describe
%% the work being presented. Separate the keywords with commas.

%%
%% This command processes the author and affiliation and title
%% information and builds the first part of the formatted document.
\maketitle

\section{Introduction}
\textbf{Textual Representations of Information:} \textit{A picture is worth a thousand words - isn't it? And hence graphical representation is by its nature universally superior to text - isn't it? Why then isn't the anecdote itself represented graphically?} - Petre \cite{petre:95}, in his advocacy for textual representation of information, challenges the notion that graphical representations of information are inherently more memorable, comprehensible, and accessible than their textual counterparts. Gershon and Page \cite{gershon:01} note that the transformation of information from a textual to visual domain, in certain instances, requires further addition of information rendering textual representations more economical. Similarly, \textit{knowing where to look} may not be obvious in visual representations of information - as validated through reading comprehension experiments \cite{green:91, green:92, blu:93} where participants were significantly slower in interpreting visual representations of the nested conditional structures within a program compared to their textual representations. This being said, these studies do not intend to dissuade the use of visual representations but rather establish the importance of textual representation of information. Often, the interplay of these paradigms brings out the best of both \cite{segel:10}. Thus, having established the importance of textual representations of information, we next explore how these notions tie into Data-to-text (D2T) generation.
\begin{figure}[h]
    \centering
    \includegraphics[scale=0.26]{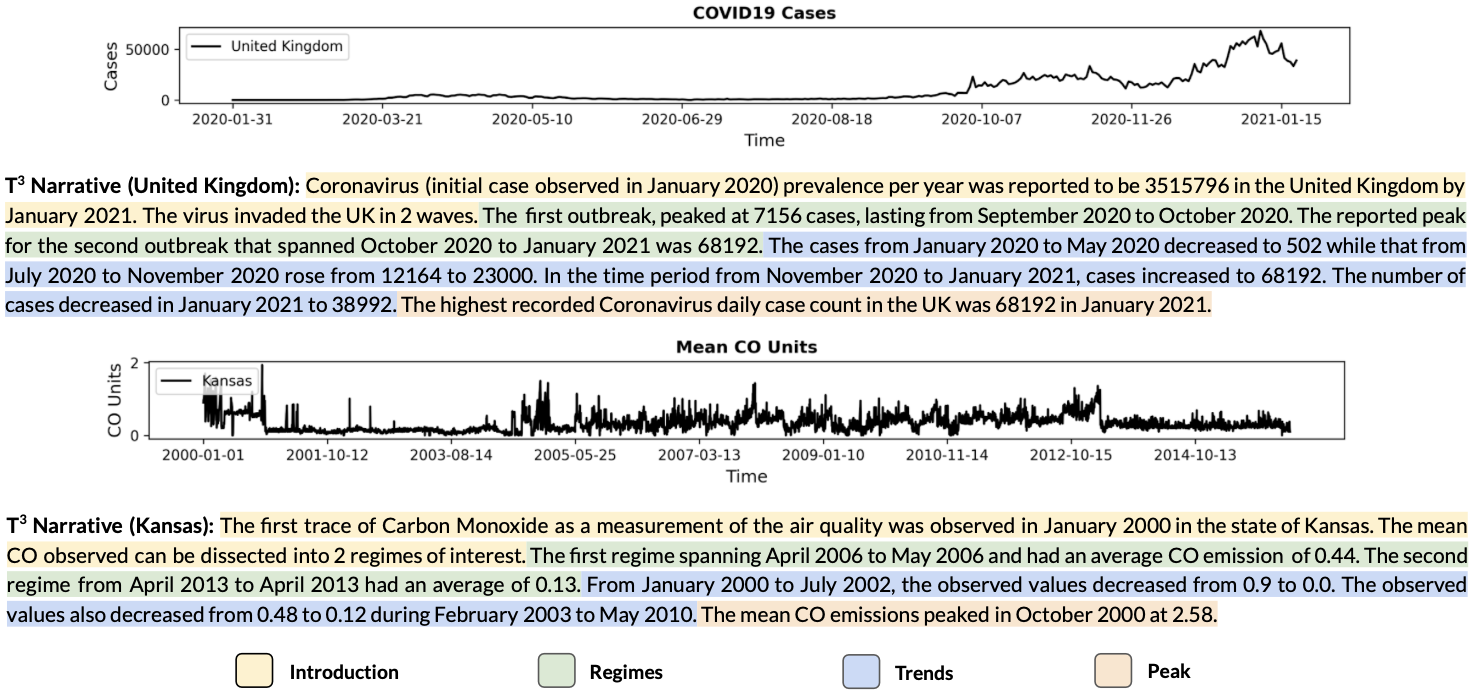}
    \caption{\footnotesize{Illustration of D2T: Narration of time-series data (COVID19 progression in the United Kingdom at the top, the Carbon-Monoxide emissions in the state of Kansas, United States at the bottom) with the LLM-based framework $T^{3}$ (T-Cube) \cite{sharma:21}. This D2T framework consumes a time-series as input and generates narratives that highlight the progression and points-of-interest (regimes, trends, and peaks) in the data through LLM-generated narratives.}}
    \label{fig:tcube}
\end{figure} 

\subsection{Defining Data-to-text Generation and Scope of the Survey}
Textual representations of information, for easier assimilation, are often presented as annotations outlining different behaviours of the underlying data stitched together. These stitched annotations, as showcased in Fig. \ref{fig:tcube}, are referred to as \textit{narratives}. The automated generation of such narratives, although serving several niches (see below), are most prevalent in the public eye through the practice of robo-journalism \cite{leppanen:17, diakopoulos:19}. Bloomberg News generates a third of its content with Cyborg, their in-house automation system that can dissect tedious financial reports and churn out news articles within seconds \cite{robo:19}. Also prevalent are the use of such systems in business intelligence (BI) settings with prominent commercial frameworks\footnote{This survey exclusively focuses on academic innovations for data-to-text generation as the technologies underlying commercial frameworks are often proprietary.} such as Arria NLG\footnote{\url{https://www.arria.com/}}, Narrative Science\footnote{\url{https://narrativescience.com/}}, and Automated Insights\footnote{\url{https://automatedinsights.com/}}.

The practice of automating the translation of data to user-consumable narratives through such systems is known as \textit{data-to-text generation}, as depicted in Fig. \ref{fig:tcube}. Although encompassed by the general umbrella of \textit{Natural Language Generation}, the nuance that differentiates D2T from the rest of the NLG landscape is that the input to the system has to qualify as a \textit{data} instance. Reiter and Dale (1997) \cite{reiterdale:97} describe the instance as a non-linguistic representation of information, and although narration of images and videos \cite{dong:21} has garnered interest in the NLG community, the definition of data-to-text employed by this survey follows that established by the seminal works prior \cite{reiterdale:97, survey:18}: an entity that is \textit{not exclusively linguistic} - tabular databases, graphs and knowledge bases, time-series, and charts. Using this clause, we limit the scope of our analysis and exclude examination of all other NLG systems that either both ingest and expel linguistic entities for downstream tasks such as machine translation \cite{jones:12,gnmt:16} or summarization \cite{nenkova:11, liu:15} or ingest non-conventional data such as images \cite{image} and videos \cite{video}. 

\vspace{0.25cm}
\par\noindent Outside of dataset specific tasks, practical applications of D2T include, but are not limited to:
\begin{itemize}
    \item Weather forecasts \cite{reiter:05, belz:08}
    \item Sport summaries \cite{robin:95, tanaka:98, barzilay:05}
    \item Healthcare \cite{portet:09, pauws:19} 
    \item Virtual dietitians \cite{anselma:18}
    \item Stock market comments \cite{murakami:17, aoki:18}
    \item Video-game dialouges \cite{juraska:19} and Driving feedback \cite{braun:18} 
\end{itemize}
 With our scope defined, below we outline the rationale for this survey, followed by a structured examination of approaches, benchmark datasets, and evaluation protocols that constitute the D2T landscape with the intent to outline promising avenues for further research.

\subsection{Survey Rationale}
Following the seminal work by Reiter and Dale \cite{reiterdale:97}, the most comprehensive survey on D2T to-date has been that by Gatt and Krahmer \cite{survey:18}. Although several articles have taken a close examination of NLG sub-fields such as dialogue systems \cite{santhanam:19}, poetry generation \cite{oliveira:17}, persuasive text generation \cite{duerr:21}, social robotics \cite{foster:19}, or exclusively focus on issues central to NLG such as faithfulness \cite{li:22} and hallucination \cite{ji:22}, a detailed break-down of the last half-decade of innovations has been missing since the last exhaustive body of work. The need for a close and consolidated examination of developments in neural D2T is more pertinent now than ever. Further, D2T distinguishes itself from other NLG tasks as it blends the generation of narratives with numerical reasoning between data points. Outside of the D2T niche, there are research communities focused on solving these individual problems - NLG \cite{chatgpt, llama, survey:18} and numerical reasoning \cite{num:1,num:2,num:3,wolfram:2023}. Thus, neural D2T is uniquely positioned such that it either has to incorporate innovations from these seemingly disparate niches or jointly innovate on both fronts. We believe this provides added justification for D2T requiring its own comprehensive literature review.

As such, neural D2T borrows heavily from advances in other facets of NLG such as \textit{neural machine translation} (NMT) \cite{bahdanau:15, gnmt:16} and \textit{spoken dialogue systems} (SDS) \cite{wen:15,wen:16,duvsek:16}. As such, the pertinence of such a survey also spans highlighting the stages of technological adoptions in the D2T paradigm and drawing distinctions between its NMT and SDS neighbors. Further, the adoption of such technologies brings about the adoption of shared pitfalls - inconsistencies in evaluation metrics \cite{reiter:18} and meaningful inter-model comparisons \cite{reimers:17}. Thus, in addition to an exhaustive examination of neural D2T frameworks, a consolidated resource on approaches to its evaluation is also necessary. Also crucial, is the discussion of benchmark datasets across shared tasks. The above considerations motivate our survey on the neural D2T paradigm intended to serve the following goals:
\begin{itemize}
    \item Structured examination of innovations in neural D2T in the last half-decade spanning relevant frameworks, datasets, and evaluation measures.
    \item Outlining the technological adoptions in D2T from within and outside of the greater NLG umbrella with the distinctions and shared pitfalls that lie therein.
    \item Highlighting promising avenues for further D2T research and exploration that promote fairness and accountability along with linguistic prowess.
\end{itemize}

\section{Datasets for Data-to-text Generation}
The first set of technological adoptions from NLP takes the form of \textit{dataset design}: parallel corpora that align the data to their respective narratives are crucial for end-to-end learning, analogous to any neural-based approach to text processing. The initial push towards building such datasets began with database-text pairs of weather forecasts \cite{reiter:05,belz:08} and sport summaries \cite{barzilay:05}. These datasets, and the convention that currently follows, use \textit{semi-structured} data that deviates from the raw numeric signals initially used for D2T systems \cite{reiter:07}. The statistics for prominent datasets among the ones discussed below are detailed in Table \ref{table:dataset}.

\subsection{Meaning Representations}
Mooney \cite{mooney:07} defines a \textit{meaning representation language} (MRL) as a formal unambigious language that allows for automated inference and processing wherein natural language is mapped to its respective \textit{meaning representation} (MR) through semantic parsing \cite{ge:05}. Robocup \cite{robocup:08}, among pioneering MR-to-text datasets, offers data from 1539 pairs of temporally ordered simulated soccer games in the form of MRs (pass, kick, turnover) accompanied with their respective human commentation. In order to mitigate the cost of building large-scale MR datasets, Liang \textit{et. al.} \cite{liang:09} use grounded language acquisition to construct WeatherGov - a weather forecasting dataset with 29528 MR-text pairs, each consisting of 36 different weather states. \textit{Abstract} meaning representation (AMR) \cite{banarescu:13}, similarly, is a linguistically grounded semantic formalism representing the meaning of a sentence as a directed graph, as depicted in Fig. \ref{fig:graphs}a. The LDC repository\footnote{\url{https://amr.isi.edu/}} hosts various AMR-based corpora. Following this, using simulated dialogues between their statistical dialogue manager \cite{young:10} and an agenda-based user simulator \cite{schatzmann:07}, Mairesse \textit{et. al.} \cite{bagel:10} offer BAGEL - an MR-text collection of 202 Cambridge-based restaurant descriptions each accompanied with two \textit{inform} and \textit{reject} dialogue types. Wen \textit{et. al.} \cite{wen:15}, through crowdsourcing, offer an enriched dataset conisiting of 5192 instances of 6 additional dialogue act types such as \textit{confirm} and \textit{informonly} (8 total) for hotels and restaurants in San Francisco. Novikova \textit{et. al.} \cite{novikova:16} show that crowdsourcing with the aid of pictorial stimuli yeild better phrased references compared to textual MRs. Following this, they released the E2E dataset\footnote{\url{http://www.macs.hw.ac.uk/InteractionLab/E2E/}} as a part of the E2E challenge \cite{e2e:17}. With 50,602 instances of MR-text pairs of restaurant descriptions, its lexical richness and syntactic complexity provides new challenges for D2T systems. Table \ref{table:mrs} showcases comparative snapshots of the aforementioned datasets.

\begin{table}[h]
\scriptsize
\centering
\caption{\footnotesize{Comparative showcase of sample MRs (and their corresponding narratives) from the RoboCup, WeatherGov, BAGEL, SF Hotels and Restaurants, and E2E datasets.}}
\label{table:mrs}
\begin{tabular}{ll}
\toprule
\multicolumn{1}{c}{MR}& \multicolumn{1}{c}{Text}\\ \hline
\begin{tabular}[c]{@{}l@{}}\textbf{RoboCup}\cite{robocup:08}\\ badPass(arg1=pink11,...), ballstopped()\\ ballstopped(), kick(arg1=pink11)\\ turnover(arg1=pink11,...)\end{tabular}                                                                                           & \begin{tabular}[c]{@{}l@{}}pink11 makes a bad pass and was picked off by purple3\end{tabular}                                                                                                                                      \\ \hline
\begin{tabular}[c]{@{}l@{}}\textbf{WeatherGov}\cite{liang:09}\\ rainChance(time=26-30,...), temperature(time=17-30,...)\\ windDir(time=17-30,...), windSpeed(time=17-30,...)\\ precipPotential(time=17-30,...), rainChance(time=17-30,...)\end{tabular} & \begin{tabular}[c]{@{}l@{}}Occasional rain after 3am. Low around 43. South wind between 11 and \\ 14 mph. Chance of precipitation is 80\%. New rainfall amounts between \\ a quarter and half of an inch possible.\end{tabular} \\ \hline
\begin{tabular}[c]{@{}l@{}}\textbf{BAGEL} \cite{bagel:10}\\ inform( name(the Fountain) \\ near(the Arts Picture House) \\ area(centre), pricerange(cheap))\end{tabular}                                                                                                                  & \begin{tabular}[c]{@{}l@{}}There is an inexpensive restaurant called the Fountain in the centre of \\ town near the Arts Picture House\end{tabular}                                                                                \\ \hline
\begin{tabular}[c]{@{}l@{}}\textbf{SF Hotels \& Rest.} \cite{wen:15}\\ inform( name=”red door cafe”, \\ goodformeal=”breakfast”, \\ area=”cathedral hill”, kidsallowed=”no”)\end{tabular}                                                                                       & \begin{tabular}[c]{@{}l@{}}red door cafe is a good restaurant for breakfast in the area of cathedral \\ hill and does not allow children.\end{tabular}                                                                             \\ \hline
\begin{tabular}[c]{@{}l@{}}\textbf{E2E} \cite{novikova:16}\\ name{[}Loch Fyne{]}, eatType{[}restaurant{]},\\ food{[}French{]}, priceRange{[}less than £20{]},\\ familyFriendly{[}yes{]}\end{tabular}                                                                                           & \begin{tabular}[c]{@{}l@{}}Loch Fyne is a family-friendly restaurant providing wine and cheese \\ at a low cost.\end{tabular}                                                                                                        \\ \bottomrule
\end{tabular}
\end{table}

\subsection{Graph Representations}
\begin{figure}[h]
    \centering
    \subfloat[\scriptsize \centering AMR]{{\includegraphics[scale=0.22]{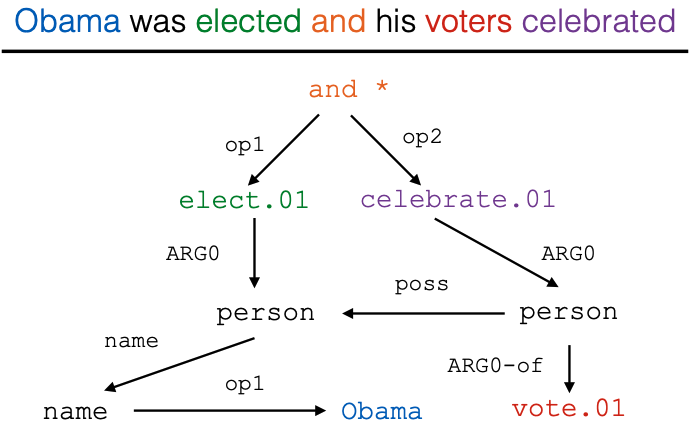} }}%
    \subfloat[\scriptsize \centering Knowledge Graph]{{\includegraphics[scale=0.15]{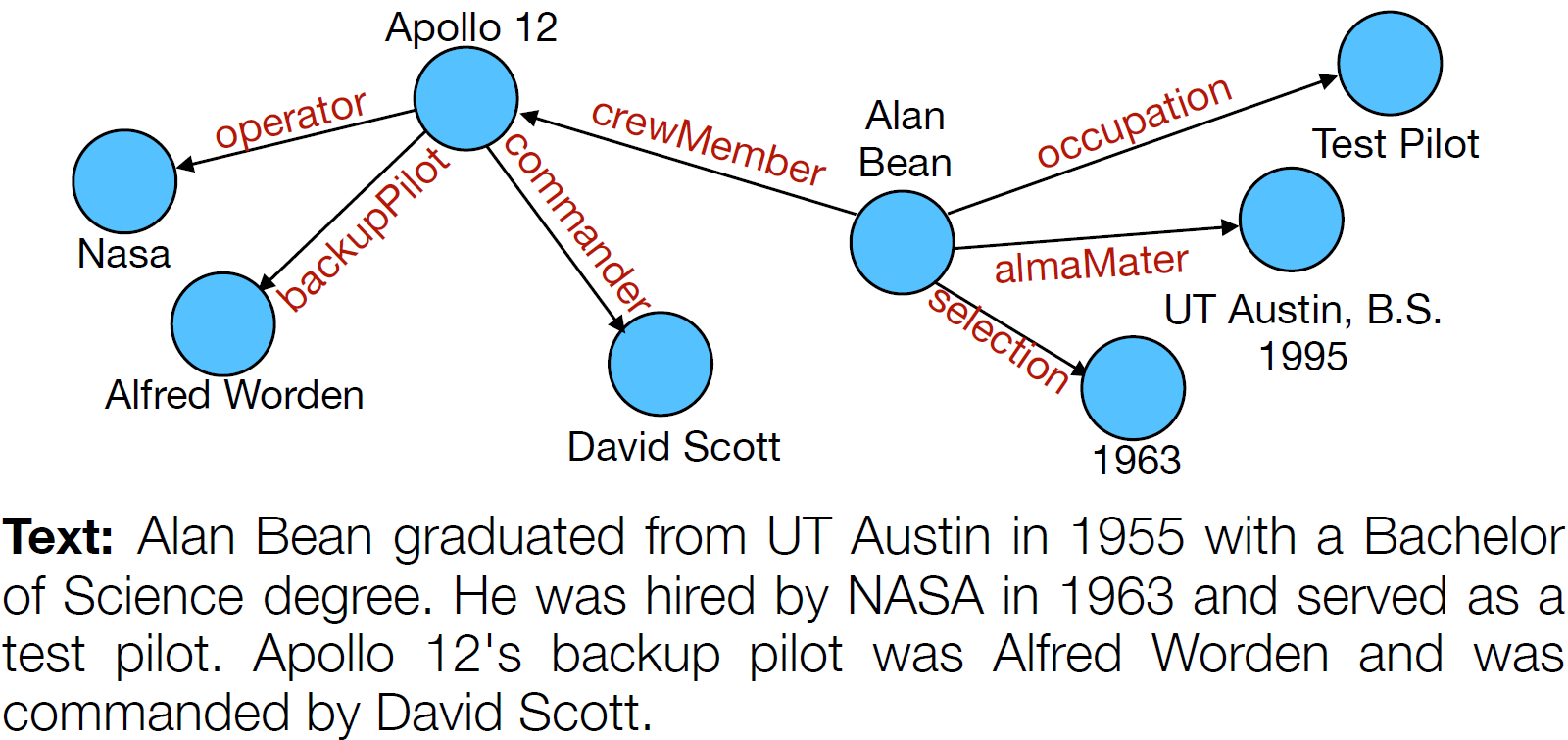} }}%
    \caption{\footnotesize AMR \cite{konstas:17} and knowledge graph \cite{ribeiro:21} snapshots, representing variants of graph-based inputs to D2T systems.}%
    \label{fig:graphs}
\end{figure}

Graph-to-text translation is not only central to D2T as its application carries over to numerous NLG fields such as question answering \cite{he:17, duan:17}, summarization \cite{fan:19}, and dialogue generation \cite{liu_b:18, moon:19}. Further, the D2T frameworks for graph-to-text borrow heavily from the theoretic formulations offered from the literature in the field of graph neural networks (GNNs), as will be discussed in \textsection 5.1.5. The \textit{domain-specific} benchmark datasets, as discussed above (see \textsection 2.1) inherently train models to generate stereotypical domain-specific text. By crowdsourcing annotations for DBPedia \cite{mendes:12} graphs spanning 15 domains, Gardent \textit{et. al.} \cite{gardent:17} introduce the WebNLG dataset\footnote{\url{https://webnlg-challenge.loria.fr/}}. The data instances are encoded as \textit{Resource Description Format} (RDF) triples of the form \textit{(subject, property, object)} as depicted in Fig. \ref{fig:graphs}b - \textit{(Apollo 12, operator, NASA)}. With 27,731 multi-domain graph-text pairs, WebNLG offers more semantic and linguistic diversity than previous datasets twice its size \cite{wen:16}. The abstract generation dataset (AGENDA) \cite{koncel:19}, built with knowledge graphs extracted from articles in the proceedings of AI conferences \cite{ammar:18} using SciIE \cite{luan:18}, offers 40,000 graph-text pairs of the article abstracts. To further promote generation challenges and cross-domain generalization, Nan \textit{et. al.} \cite{nan:21} merge the E2E and WebNLG dataset with large heterogeneous collections of diverse predicates from Wikipedia tables annotated with tree ontologies to generate the data-record-to-text (DART) corpus. With 82,191 samples, this resulting open-domain corpus is almost quadruple the size of WebNLG.

\subsection{Tabular Representations}
\begin{figure}[h]
    \centering
    \includegraphics[scale=0.22]{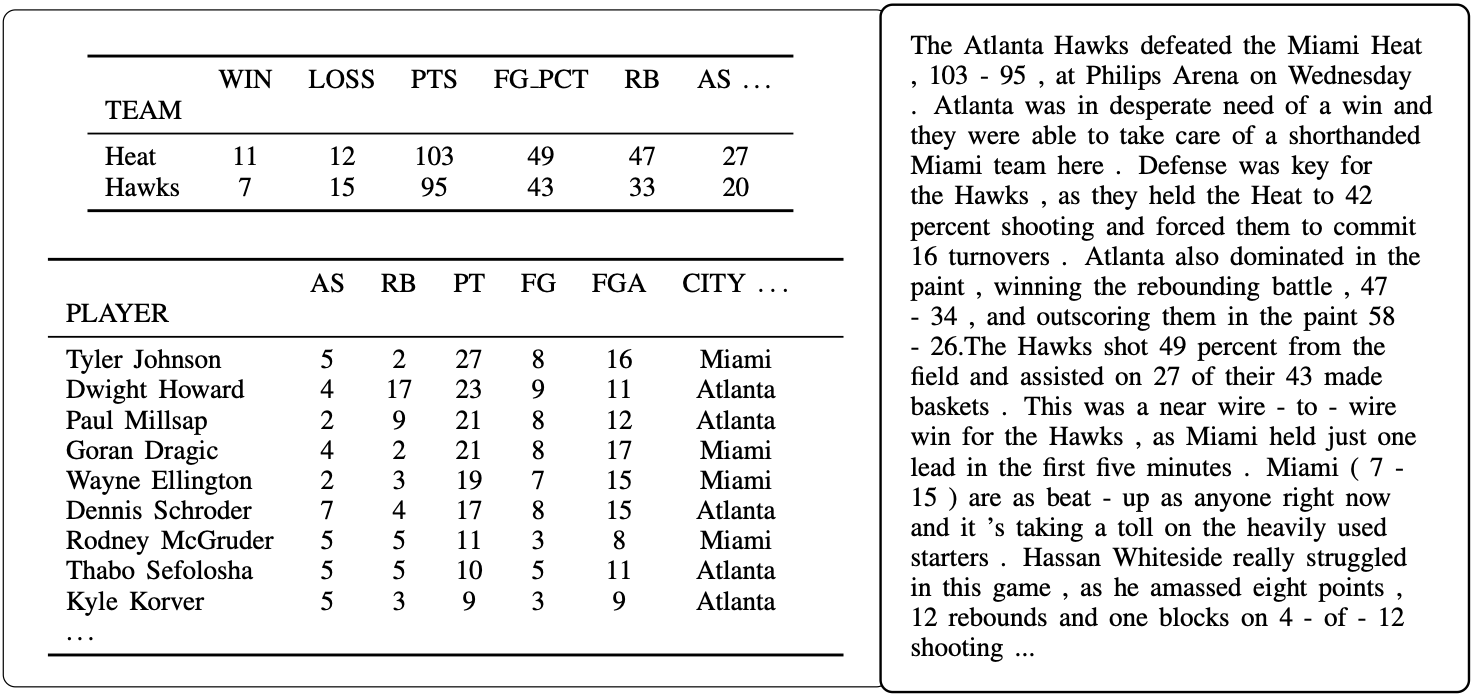}
    \caption{\footnotesize Showcasing the intent of T2T, the statistics of a basketball match between the Atlanta Hawks and the Miami Heat (left) is to be translated into easily consumable narratives (right). Snapshot from the RotoWire dataset \cite{wiseman:17}.}
    \label{fig:roto}
\end{figure}

Information represented in large tables can be difficult to comprehend at a glance, thus, table-to-text (T2T) aims to generate narratives highlighting crucial elements of a tabular data instance through summarization and logical inference over the table - as showcased in Figure \ref{fig:roto}. Similar to graph-to-text, the underpinnings of tabular representation learning is also shared with other fields outside of NLG, such as the generation of synthetic network traffic \cite{network:1, network:2}.

WikiBio \cite{lebret:16}, as an initial foray towards a large-scale T2T dataset, offers 700k table-text pairs of Wikipedia info-boxes with the first paragraph of its associated article as the narrative. With a vocabulary of 400k tokens and 700k instances, WikiBio offers a substantially larger benchmark compared to the pioneering WeatherGov and Robocup datasets that have less than 30k data-text pairs. For neural systems, as the length of output sequence increases, the generated summary diverges from the reference. As such, the RotoWire dataset \cite{wiseman:17} (Fig. \ref{fig:roto}), consisting of verbose descriptions of NBA game statistics, brings forth new challenges in long-form narrative generation as the average reference length of RotoWire is 337 words compared to 28.7 of WikiBio. Similarly, with the observation that only 60\% of the content in RotoWire narratives can be traced back to the data records, Wang \cite{wang:19} introduce RotoWire-FG, a refined version of the original dataset aimed at tackling divergence (see \textsection 3.2), where narrative instances not grounded by their respective tables are removed from the dataset.
TabFact \cite{chen_d:20} contains annotated sentences that are either supported or refuted by the tables extracted from Wikipedia. Similar to RotoWire-FG, Chen \textit{et. al.} \cite{chen_c:20} offer a filtered version of TabFact by retaining only those narratives that can be logically inferred from the table.

For controlled generation, Parikh \textit{et. al.} \cite{parikh:20} propose ToTTo which generates a single sentence description of a table on the basis of a set of highlighted cells where annotators ensure that the target summary only contains the specified subset of information. With over 120k training samples, ToTTo establishes an open-domain challenge for D2T in controlled settings. Similarly, to evaluate narrative generation in open domain settings with sentences that can be logically inferred from mathematical operations over the input table, Chen \textit{et. al.} \cite{chen_c:20} modify the reference narratives of the TabFact dataset to construct LogicNLG with 7392 tables. Following this, with tables and their corresponding descriptions extracted from scientific articles, Moosavi \textit{et. al.} \cite{moosavi:21} introduce SciGen, where the narratives include arithmetic reasoning over the tabular numeric entries. Building upon the long form generation premise of RotoWire, Chen \textit{et. al.} \cite{chen_b:21} construct WikiTableT, a multi-domain table-text dataset with 1.5 million instances pairing Wikipedia descriptions to their corresponding info-boxes along with additional hyperlinks, named-entities, and article metadata. The majority of these datasets are available in a unified framework through TabGenie\footnote{\url{https://pypi.org/project/tabgenie/}}.

\begin{table}[h]
\footnotesize
\centering
\caption{\footnotesize{Highlights from prominent D2T datasets: format, number of samples (size), the number of linguistic tokens across the dataset (tokens), and availability of non-anglo-centric variants.}}
\label{table:dataset}
\begin{tabular}{@{}llll@{}}
\toprule
\multicolumn{1}{c}{Benchmark} & \multicolumn{1}{c}{Format} & \multicolumn{1}{c}{Size} & \multicolumn{1}{c}{Tokens} \\ \midrule
E2E                      & MR                         & 50,602                   & 65,710                     \\
LDC2017T10               & AMR                        & 39,260                   & \multicolumn{1}{c}{-}      \\
WebNLG (en, ru)               & RDF                        & 27,731                   & 8,886                      \\
DART                          & RDF                        & 82,191                   & 33,200                     \\
WikiBio                       & Record                     & 728,321                  & 400,000                    \\
RotoWire                      & Record                     & 4,853                    & 11,300                     \\
TabFact                       & Record                     & 16,573                   & \multicolumn{1}{c}{-}      \\
ToTTo                         & Record                     & 120,000                  & 136,777                    \\
LogicNLG                         & Record                     & 37,000                  & 52,700                    \\ 
WikiTableT                          & Record                        & 1.5M                   & 169M                     \\
\bottomrule
\end{tabular}
\end{table}

\subsection{Data Collection \& Enrichment}
The majority of the prominent datasets discussed in \textsection 2.1 - \textsection 2.3 are either collected by merging aligned data-narrative pairs that occur naturally in the \say{wild} \cite{lebret:16, wiseman:17} or are collected through dedicated crowd-sourcing approaches \cite{e2e:17, gardent:17}. However, there are notable works that employ a hybrid approach to data collection. CACAPO \cite{cacapo:20}, an MR-style multi-domain dataset, follows a collection process inspired by Oraby \textit{et. al.} \cite{oraby:19} wherein the naturally-occurring narratives are first scraped from the internet and are later manually annotated to generate attribute-value pairs. Similarly, Chart-to-Text \cite{kanthara:22} follows a similar mechanism of data collection wherein candidate narratives for each chart are first automatically generated via a heuristic-based approach and then are rated by crowd-sourced workers. In similar lines, the ToTTo dataset \cite{parikh:20} discussed in \textsection 2.3 uses crowd-sourced annotators as data \say{cleaners} - iteratively improving upon the automatically scraped narratives, rather than annotating them from scratch - thus greatly reducing the cost of data acquisition. In addition to innovations in data collection, efforts from the D2T community has also focused on the enrichment of existing datasets. As such, Ferreira \textit{et. al.} \cite{e-webnlg} augment the WebNLG dataset with intermediate representation for discourse ordering and referring expression generation. By manually delexicalizing  (see \textsection 4.1) the narratives, Ferreira \textit{et. al.} were able to automatically extract a collection of referring expressions by tokenizing the original and delexicalized texts and finding the non-overlapping tokens between them. Similarly, the authors also extracted the order of the arguments in the text by referring to the order of the general tags in the delexicalized texts. This work has also been extended to enrich the E2E dataset \cite{e-e2e}.

\section{Data-to-text Generation Fundamentals and Notations}
\subsection{What to Say and How to Say It}
\begin{figure*}[t]
    \centering
    \includegraphics[scale=0.32]{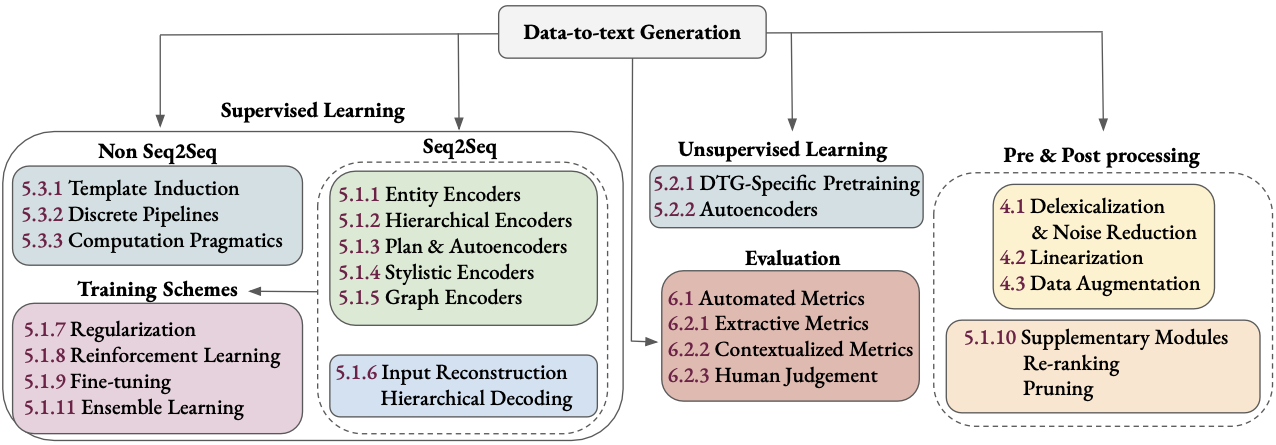}
    \caption{\footnotesize{Data-to-text Generation Taxonomy Corresponding to Sections in the Survey Design}}%
    \label{fig:main}
\end{figure*}

The data instance, typically, contains more information than what we would intend for the resulting narrative to convey - verbose narratives that detail every attribute of the data instance contradicts the premise of consolidation. Thus, to figure out \textit{what to say}, a subset of the original information content is filtered out based on the target audience through the process of \textit{content selection}. Starting from data-driven approaches such as clustering \cite{duboue:03} and the use of hidden Markov models (HMMs) \cite{barzilay:04}, the attention of the research community has recently shifted to learning \textit{alignments} between the data instance and its narrative \cite{liang:09}. Bisazza and Marcello \cite{bisazza:16} note that pre-reordering the source words to better resemble the target narrative yeilds significant improvements in NMT.  Prior to neural explorations, learning this alignment has been explored with log-linear models \cite{angeli:10} and tree representations \cite{konstas:12, konstas:13}. With \textit{what to say} determined, the next step lies in figuring out \textit{how to say it}, that is, the construction of words, phrases, and paragraphs - this realization of the narrative structure is known as \textit{surface realization}. While traditionally, the processes of content selection and surface realization \cite{reiterdale:97, jurafsky:14} act as discrete parts of the generation pipeline, the neural sequence-to-sequence (seq2seq) paradigm jointly learns these aspects. For a peripheral view of the articles discussed in this section, Table \ref{table} highlights prominent papers categorized based on their D2T tasks and the benchmark datasets used. Similarly, Fig. \ref{fig:main} outlines the organization of the remainder of this survey.

\subsection{Hallucinations and Omissions}
Apart from the importance of coherence and linguistic diversity in surface realization, \textit{data fidelity} is a crucial aspect of D2T systems - the narrative should neither \textit{hallucinate} contents absent from the data instance nor \textit{omit} contents present in the data instance. Often, the divergence present in benchmark training datasets, wherein the narrative may contain data absent from the source or not cover the entirety of the data instance, is the culprit behind hallucination tendencies in the model \cite{rohrbach:18}. Often times, the need for both linguistic diversity and data fidelity turns into a balancing act between conflicting optimization objectives leading to novel challenges \cite{hashimoto:19}. While almost all of the D2T approaches discussed below engage in balancing coherence and diversity with data-fidelity (besides \textsection 5.1.4 Stylistic Encoding), overarchingly, the approach to balancing these conflicting objectives can be thought to take place in two forms:
\begin{itemize}
    \item \textbf{Architectural Interventions}: The sections \textsection 5.1.1 Entity Encoders, \textsection 5.1.2 Hierarchical Encoders, \textsection 5.1.3 Plan Encoders \& Autoencoders, \textsection 5.1.5 Graph Encoders, \textsection 5.1.6 Reconstruction \& Hierarchical Decoders, and \textsection 5.1.10 Supplemental Frameworks suggest modifications/augmentations to the seq2seq architecture such that it fosters data-fidelity tendencies.
    \item \textbf{Loss-function Interventions}: An alternative avenue to achieving a balance between conflicting optimization objectives is to directly model the objective functions to perform multi-task learning: as such sections \textsection 5.1.7 Regularization Techniques and \textsection 5.1.8 Reinforcement Learning suggest modifications/augmentations to the seq2seq loss functions.
\end{itemize}

\subsection{Establishing Notation and Revisiting Seq2Seq}
\begin{figure}[h]
    \centering
    \includegraphics[scale=0.3]{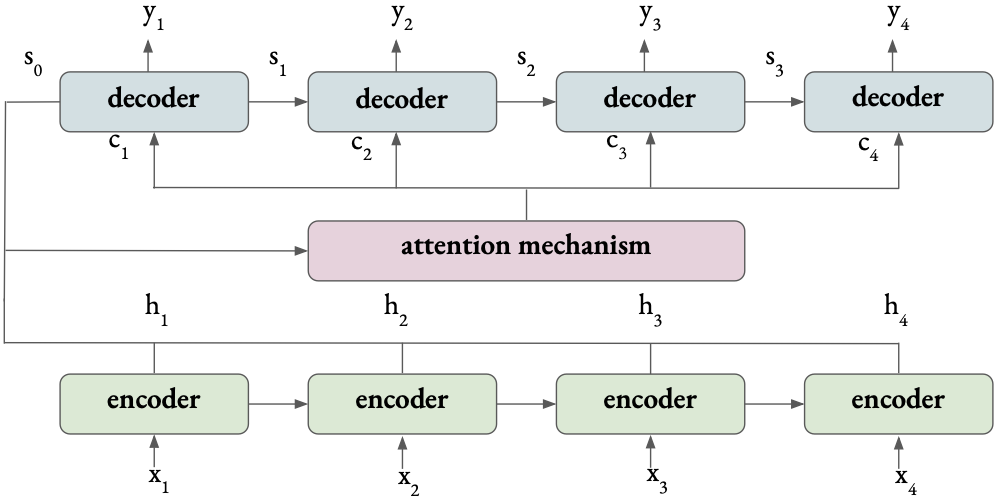}
    \caption{\footnotesize{Attention-based seq2seq framework: The encoder consumes the sequential input translating it to a weighed hidden representation to be then consumed and decoded into linguistic tokens by the decoder.}}
    \label{fig:seq2seq}
\end{figure}

For the consistency and readability of this survey, the notation outlining the basic encoder-decoder seq2seq paradigm \cite{sutskever:11, cho:14, bahdanau:15, vaswani:17} in D2T (Fig. \ref{fig:seq2seq}), as defined below and respectively compiled in Table \ref{table:notations}, will remain valid throughout unless stated otherwise. However, the namespace for additional variable definitions in the individual sections will be limited to their mentions. Let $S = \{x_{j},y_{j}\}_{j=1}^{N}$ be a dataset of $N$ data instances $x$ accompanied with its natural language narrative $y$. Based on the construction of $S$, $x$ can be a set of $K$ data records $x = \{r_{j}\}_{j=1}^{K}$ with each entry $r$ comprised of its respective entity $r.e$ and value $r.m$ attributes or $x$ can be an instance of a directed graph $x = (V,E)$ with vertices $v \in V$ and edges $(u,v) \in E$. In the RotoWire instance (Figure \ref{fig:roto}), for $r_{j}$ = Heat, $r_{j}.e$ = WIN attribute  would have value $r_{j}.m$ = 11. Given pairs $(x,y)$, the seq2seq model $f_{\theta}$ is trained end-to-end to maximize the conditional probability of generation $P(y|x)=\prod_{t=1}^{T} P(y_{t}|y_{<t}, x)$. The parameterization of $f_{\theta}$ is usually carried out through RNNs such as LSTMs \cite{bengio:94, hochreiter:97} and GRUs \cite{cho:14}, or transfomer\footnote{Though the base transformer architecture is oblivious to input structures, we assume positionally encoded transformers to fall into the seq2seq paradigm.} architectures \cite{vaswani:17}. For attention-based RNN architectures with hidden states $h_{t}$ and $s_{t}$ for the encoder and decoder respectively, the context vector $c_{t}= \sum_{i} \alpha_{t,i} h_{i}$ weighs the encoder hidden states with attention weights $\alpha_{t,i}$. While Bahdanau \textit{et. al.} \cite{bahdanau:15} use a multi-layer perceptron (MLP) to model $\alpha_{t,i}$, several alterations to modeling the attention weights have been proposed \cite{luong:15, vinyals:15}.

For handling of out-of-vocabulary (OOV) tokens, Gu \textit{et. al.} \cite{gu:16} attempt to model the rote memorization process of human learning where a language model conditioned on binary variable $z_{t} \in \{0,1\}$ can either generate $p_{gen}$ the next token or copy it from the source $p_{copy}$ based on their respective probabilities. While Gu \textit{et. al.} \cite{gu:16} and Yang \textit{et. al.} \cite{yang:17} parameterize the \textit{joint} distribution over $y_{t}$ and $z_{t}$ directly (\ref{eq:2}), Gul{\c{c}}ehre \textit{et. al.} \cite{gul:16} decompose the joint probability (\ref{eq:3}), using an MLP to model $p(z_{t} | y_{<t}, x)$. 
\begin{align}
    P(y_{t},z_{t}|y_{<t},x) \propto
    \begin{cases}
    \label{eq:2}
        p_{gen}(y_{t},y_{<t}, x) \: \: z_{t} = 0  \\
        p_{copy}(y_{t},y_{<t}, x) \: z_{t} = 1, y_{t} \in x \\
        0  \qquad \qquad \qquad \: z_{t} = 1, y_{t} \notin x
    \end{cases} \\
    \begin{cases}
    \label{eq:3}
         p_{gen}(y_{t}|z_{t},y_{<t}, x) \: p(z_{t}|y_{<t}, x)  \: z_{t} = 0 \\
         p_{copy}(y_{t}|z_{t},y_{<t}, x) \: p(z_{t}|y_{<t}, x) \: z_{t} = 1
    \end{cases}
\end{align}
Similar to the greater NLG paradigm, different strategies for modeling the conditional probability of generation $P(y|x)$, the attention mechanisms $\{\alpha_{t,i},c_{t}\}$, and the copy mechanisms $\{p_{gen},p_{copy}\}$, as discussed below, \textit{often form the basis for D2T innovations}. In addition to this, variations in training strategies such as teacher-forcing \cite{teacher}, reinforcement learning \cite{reinforcement}, and autoencoder-based reconstruction \cite{recon} open up further avenues for D2T innovation.

\begin{table}[h]
\footnotesize
\centering
\caption{\footnotesize{Notation descriptions}}
\label{table:notations}
\begin{tabular}{@{}ll@{}}
\toprule
Notation & Description \\ \midrule
$S$        & Dataset             \\
$(x,y) \in S$         & Data instance $x$ and its natural language representation $y$             \\
$r = (r.e, r.m)$         & Data record $r$ with its entity $r.e$ and value $r.m$ attributes             \\
$G = (V,E)$         &  Graph instance $G$ with vertices $V$ and edges $E$           \\
$(u,v) \in E$         &  Nodes $u$ and $v$ of an edge $E$           \\
$f_{\theta} \in \{f_{1},...,f_{n}\}$         &  Model $f_{\theta}$ that may belong to an ensemble $\{f_{1},...,f_{n}\}$         \\
$P(y|x)$         &    Conditional probability of sequence $y$ given $x$         \\
$h_{t}, s_{t}$         &  Encoder $h_{t}$ and decoder $s_{t}$ hidden state representations           \\
$c_{t},\alpha_{t,i}$         &  Context vector $c_{t}$ weighing $h_{t}$ with attention weights $\alpha_{t,i}$ \\
$p_{gen}, p_{copy}$         &   Token generation $p_{gen}$ or copying $p_{copy}$ probabilities \\ 
$z_{t}\in \{0,1\}$         & Binary variable that selects either $p_{gen}$ or $p_{copy}$            \\ 
$W_{i \in \mathbb{N}}$, $b_{i \in \mathbb{N}}$    &   Arbitrary weights and biases parameterizing  $f_{\theta}$ \\ 
\bottomrule
\end{tabular}
\end{table}

\section{Innovations in Data Preprocessing}
Contrary to the other facets of NLG, such as chatbots, for which large-scale data can be harvested \cite{lowe:15,abbott:16}, D2T datasets are often smaller in scale and task-specific. Ferreira \textit{et. al.} \cite{ferreira:17} note that phrase-based translation models \cite{koehn:07} can outperform neural models in such data sparsity. As such, \textit{delexicalization}, \textit{noise reduction}, \textit{linearization}, and \textit{data augmentation} are preprocessing techniques often employed to tackle said sparsity of training data.

\subsection{Delexicalization \& Noise Reduction}
\begin{figure}[h]
    \centering
    \subfloat[\scriptsize \centering Delexicalization in MRs]{{\includegraphics[width=6.5cm]{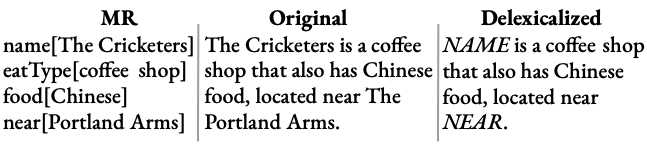} }}
    \subfloat[\scriptsize \centering Linearization of an RDF graph ]{{\includegraphics[width=6.5cm]{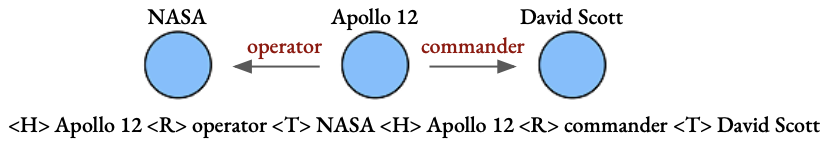} }}%
    \caption{\footnotesize Illustrations of delexicalization in MRs \cite{shimorina:18} and linearization of graphs \cite{ribeiro:21}.}%
    \label{fig:delex}
\end{figure}

\textit{Delexicalization}, often referred to as \textit{anonymization}, is a common practice in D2T \cite{bagel:10,duvsek:16} wherein the slot-value pairs for the entities and their attributes in training utterances are replaced with a placeholder token such that weights between similar utterances can be shared \cite{nayak:17} - as illustrated in Figure \ref{fig:delex}a. These placeholder tokens are later replaced with tokens copied from the input data instance \cite{lebret:16}. In comparison to copy-based methods for handling rare entities, delexicalization has shown to yield better results in constrained datasets \cite{shimorina:18}. 

From the notion that delexicalization of the data instance may cause the loss of vital information that can aid seq2seq models in sentence planning, where some data instance slots may even be deemed nondelexicalizable \cite{wen:15}, Nayak \textit{et. al.} \cite{nayak:17} explore different nondelexicalized input representations (mention representations) along with grouping representations as a form of sentence planning (plan representations). The authors note improvements over delexicalized seq2seq baselines when input mentions are concatenated with each slot-value pair representing a unique embedding. The efficacy of such concatenation is also corroborated by Freitag and Roy \cite{freitag:18}. Further, the addition of positional tokens representing intended sentence position to the input sequence offers further improvements. Addressing this, in addition to delexicalizing categorical slots, Juraska \textit{et. al.} \cite{juraska:18} employ hand-crafted tokens for values that require different treatment in their verbalization:  for the slot \textit{food}, the value \textit{Italian} is replaced by \textit{slot\_vow\_cuisine\_food} indicating that the respective utterance should start with a vowel and the value represents a cuisine - \textit{an Italian restaurant}. Perez-Beltrachini and Lapata \cite{perez:18} delexicalize numerical expressions, such as dates, using tokens created with the attribute name and position of the delexicalised token. Colin and Gardent \cite{colin:19} note performance improvements with an extensive anonymization scheme wherein all lemmatized content words (expect adverbs) are delexicalized as compared to restricting delexicalization to named entities.

The presence of narratives that fail to convey vital attributes of data instances leads to \textit{semantic noise} in the dataset \cite{howcroft:17, eric:20}. Du{\v{s}}ek \textit{et. al.} \cite{duvsek:19} employ slot matching \cite{reed:18} to clean the E2E corpus for semantic correctness and explore the impact of semantic noise on neural model performance. Similarly, Obeid and Hoque \cite{obeid:20} substitute data mentions in the narrative, identified through \textit{Named Entity Recognition} (NER), with a predefined set of tokens which are later replaced through a look-up operation. Liu \textit{et. al.} \cite{liu_b:21} focus on generating faithful narratives and truncate the reference narratives by retaining only the first few sentences since the latter are prone to have been \textit{inferred} from the table.

\subsection{Linearization}
Frameworks for MR (and graph) narration that shy from dedicated graph encoders rely on effective \textit{linearization} techniques - the representation of graphs as linear sequences, as illustrated in Figure \ref{fig:delex}b. While Ferreira \textit{et. al.} \cite{ferreira:17} note improvements in neural models with the adoption of a 2-step classifier \cite{lerner:13} that maps AMRs to the target text, Konstas \textit{et. al.} \cite{konstas:17} showcase agnosticsm to linearization orders by grouping and anonymizing graph entities for delexicalization with the Stanford NER \cite{finkel:05}. The reduction in graph complexity and subsequent mitigation of the challenge brought forth by data sparsity lends any depth-first traversal of the graph as an effective linearization approach. Moryossef \textit{et. al.} \cite{moryossef:19} append text plans modeled as ordered trees \cite{stent:04} to the WebNLG training set and use an off-the-shelf NMT system \cite{gul:16} for plan-to-text generation. However, the authors note that the restriction of requiring single entity mentions in a sentence establishes their approach as dataset dependent.

For pretrained language models such as GPT-2 \cite{radford:19} and T5 \cite{raffel:19}, Zhang \textit{et. al.} \cite{zhang_b:20} and Gong \textit{et. al.} \cite{gong:20} represent tables as a linear sequence of attribute-value pairs and use a special  token as the separator between the table data  and the reference text. It should be noted that T5, while performing the best on automated metrics, fails to generate good summaries when numerical calculations are involved. Chen \textit{et. al.} \cite{chen_c:20} traverse the table horizontally, each row at a time, where each element is represented by its corresponding field and cell value separated by the keyword \textit{is}. For scientific tables, Suadaa \textit{et. al.} \cite{suadaa:21} view a table $T_{D}$ as a set of cells with their corresponding row and column headers $h=[rh:ch]$ with $th$ for overlapping tokens, numerical value $val$, and metric-type $m$. The cells are marked with target flag $tgt$ which is set to 1 for \textit{targeted cells} and 0 otherwise respective to the content plan. The linearization of the resulting tables is done with templates that consist of concatenation $T_{D} = [h:th:val:m:tgt]$, filtration based on $tgt$, pre-computed mathematical operations, and their respective combinations.

\subsection{Data Augmentation}
Often, appending \textit{contextual} examples from outer sources to the training set, or \textit{permuting} the training samples themselves to append variation, helps mitigate data sparsity. This is known as data \textit{augmentation}. Nayak \textit{et. al.} \cite{nayak:17} propose the creation of pseudo-samples by permuting the slot orderings of the MRs while keeping the utterances intact. Juraska \textit{et. al.} \cite{juraska:18}, however, take an utterance-oriented approach where pseudo-samples are built by breaking training MRs into single-sentence utterances. For the shared surface realization task \cite{mille:18}, Elder and Hokamp \cite{elder:18} augment the training set with sentences from the WikiText corpus \cite{merity:16} parsed using UDPipe \cite{straka:17}. Following this premise, Kedzie and Mckeown \cite{kedzie:19} curate a collection of utterances from novel MRs using a vanilla seq2seq model with noise injection sampling \cite{cho:16}. The validity of the MRs associated with these utterances are computed through a CNN-based parser \cite{kim:14} and the valid entries are augmented to the training set. However, it is worth noting, that performance gains from augmenting the training set with out-of-domain (OOD) instances, tend to saturate after a certain point \cite{freitag:18}. Also, practitioners of data augmentation should note that caution is advised when augmenting with synthetic data, as the inclusion of such data may reinforce the mistakes of the model \cite{synthetic_data}.

Chen \textit{et. al.} \cite{chen:19} append knowledge-graphs representing \textit{external context} to the table-text pairs and quantify its efficacy through their metric KBGain - the ratio of tokens unique to the external context to the total number of tokens in the narrative. Similarly, Ma \textit{et. al.} \cite{ma:19} augment the limited training data for table-text pairs by assigning part-of-speech (POS) tags for each word in the reference and further increase the robustness of their model with adversarial examples created by randomly adding and removing words from the input. In contrast, Chen \textit{et. al.} \cite{chen_c:20} create adversarial examples by randomly swapping entities in the narrative with ones that appear in the table. Following this, Liu \textit{et. al.} \cite{liu_b:21} use an augmented plan consisting of table records and entities recognized from the reference narrative which eliminates the inclusion of information not present in the table. For few-shot learning, Liu \textit{et. al.} \cite{liu:21} observed that the performance of a GPT-3 model \cite{brown:20} improved upon providing in-context examples computed based on their k-nearest neighbor ($k =2$) embeddings.

\begin{table}[htp]
\scriptsize
\centering
\caption{\footnotesize{Task and dataset based summarization of noted D2T frameworks over the last half-decade.}}
\label{table}
\begin{tabular}{|llll|}
\hline
\multicolumn{1}{|c|}{Dataset}                                                                                           & \multicolumn{1}{c|}{Publication Highlights}                                    & \multicolumn{2}{c|}{Framework \& Human Evaluation}         \\ \hline
\multicolumn{4}{|c|}{\textbf{MR-to-Text}}                                                                                                                                                                                                                             \\ \hline
\multicolumn{1}{|l|}{Robocup \& WeatherGov}                                                                             & \multicolumn{1}{l|}{\citeb{mei:16} Coarse-to-fine aligner \& penalty based on learned priors} & \multicolumn{1}{l|}{LSTM $\rightarrow$ LSTM + regularization}     & N \\ \hline
\multicolumn{1}{|l|}{Recipe \& SF H\&R}                                                                                 & \multicolumn{1}{l|}{\citeb{kiddon:16} Neural agenda-checklist modeling}                          & \multicolumn{1}{l|}{GRU $\rightarrow$ GRU + agenda encoders}      & Y \\ \hline
\multicolumn{1}{|l|}{BAGEL}                                                                                             & \multicolumn{1}{l|}{\citeb{duvsek:16} Reranking beam outputs w/ RNN-based reranker}              & \multicolumn{1}{l|}{LSTM $\rightarrow$ LSTM + reranker}           & N \\ \hline
\multicolumn{1}{|l|}{Restaurant Ratings}                                                                                & \multicolumn{1}{l|}{\citeb{nayak:17} Nondelexicalized inputs w/ data augmentation}              & \multicolumn{1}{l|}{LSTM $\rightarrow$ LSTM}                      & Y \\ \hline
\multicolumn{1}{|l|}{WikiData}                                                                                          & \multicolumn{1}{l|}{\citeb{chisholm:17} Complementary text-to-data translation}                    & \multicolumn{1}{l|}{GRU $\rightarrow$ GRU}                        & Y \\ \hline
\multicolumn{1}{|c|}{\multirow{14}{*}{E2E}}                                                                             & \multicolumn{1}{l|}{\citeb{duvsek:18} Comparative evaluation of 62 systems}                      & \multicolumn{1}{l|}{Seq2Seq + Data-driven + Templated} & Y \\ \cline{2-4} 
\multicolumn{1}{|c|}{}                                                                                                  & \multicolumn{1}{l|}{\citeb{puzikov:18} MLP encoder attuned to the dataset}                        & \multicolumn{1}{l|}{MLP $\rightarrow$ GRU}                        & Y \\ \cline{2-4} 
\multicolumn{1}{|c|}{}                                                                                                  & \multicolumn{1}{l|}{\citeb{zhang:18} Two-level hierarchical encoder}                            & \multicolumn{1}{l|}{CAEncoder $\rightarrow$ GRU}                  & N \\ \cline{2-4} 
\multicolumn{1}{|c|}{}                                                                                                  & \multicolumn{1}{l|}{\citeb{juraska:18} Ensemble w/ heuristic reranking}                           & \multicolumn{1}{l|}{Ensemble w/ LSTM + CNN}            & N \\ \cline{2-4} 
\multicolumn{1}{|c|}{}                                                                                                  & \multicolumn{1}{l|}{\citeb{su:18} Hierarchical decoding with POS tags}                       & \multicolumn{1}{l|}{GRU $\rightarrow$ GRU}                        & N \\ \cline{2-4} 
\multicolumn{1}{|c|}{}                                                                                                  & \multicolumn{1}{l|}{\citeb{freitag:18} Unsupervised DTG with DAEs}                                & \multicolumn{1}{l|}{LSTM $\rightarrow$ LSTM + DAEs}               & Y \\ \cline{2-4} 
\multicolumn{1}{|c|}{}                                                                                                  & \multicolumn{1}{l|}{\citeb{gehrmann:18} Comparative evaluations w/ ensembling \& penalties}        & \multicolumn{1}{l|}{Ensemble w/ LSTM + T}              & N \\ \cline{2-4} 
\multicolumn{1}{|c|}{}                                                                                                  & \multicolumn{1}{l|}{\citeb{deriu:18} Syntactic controls with SC-LSTM}                           & \multicolumn{1}{l|}{SC-LSTM}                           & Y \\ \cline{2-4} 
\multicolumn{1}{|c|}{}                                                                                                  & \multicolumn{1}{l|}{\citeb{shen:19} Computational pragmatics based DTG}                        & \multicolumn{1}{l|}{GRU $\rightarrow$ GRU}                        & N \\ \cline{2-4} 
\multicolumn{1}{|c|}{}                                                                                                  & \multicolumn{1}{l|}{\citeb{colin:19} Extensive anonymization}                                   & \multicolumn{1}{l|}{LSTM $\rightarrow$ LSTM}                      & Y \\ \cline{2-4} 
\multicolumn{1}{|c|}{}                                                                                                  & \multicolumn{1}{l|}{\citeb{duvsek:19} Semantic correctness in neural DTG}                        & \multicolumn{1}{l|}{LSTM $\rightarrow$ LSTM}                      & Y \\ \cline{2-4} 
\multicolumn{1}{|c|}{}                                                                                                  & \multicolumn{1}{l|}{\citeb{kedzie:19} Self-training w/ noise injection sampling}                 & \multicolumn{1}{l|}{GRU $\rightarrow$ GRU}                        & Y \\ \cline{2-4} 
\multicolumn{1}{|c|}{}                                                                                                  & \multicolumn{1}{l|}{\citeb{roberti:19} Char-level GRU w/ input reconstruction}                    & \multicolumn{1}{l|}{GRU $\rightarrow$ GRU}                        & N \\ \cline{2-4} 
\multicolumn{1}{|c|}{}                                                                                                  & \multicolumn{1}{l|}{\citeb{fu:20} CRFs w/ Gumbel categorical sampling}                       & \multicolumn{1}{l|}{CRF}                               & N \\ \hline
\multicolumn{4}{|c|}{\textbf{Graph-to-Text}}                                                                                                                                                                                                                          \\ \hline
\multicolumn{1}{|l|}{AGENDA}                                                                                            & \multicolumn{1}{l|}{\citeb{koncel:19} Graph-centric Transformer \& AGENDA dataset}               & \multicolumn{1}{l|}{T $\rightarrow$ LSTM + LSTM encoding}         & Y \\ \hline
\multicolumn{1}{|l|}{LDC2015E25}                                                                                        & \multicolumn{1}{l|}{\citeb{ferreira:17} Phrase vs Neural MR-text w/ preprocessing analysis}        & \multicolumn{1}{l|}{LSTM $\rightarrow$ LSTM + Phrase-based}       & N \\ \hline
\multicolumn{1}{|l|}{\multirow{2}{*}{LDC2015E86}}                                                                       & \multicolumn{1}{l|}{\citeb{konstas:17} Unlabeled pre-training \& linearization agnosticism}       & \multicolumn{1}{l|}{LSTM $\rightarrow$ LSTM}                      & N \\ \cline{2-4} 
\multicolumn{1}{|l|}{}                                                                                                  & \multicolumn{1}{l|}{\citeb{ribeiro:19} Dual encoding for hybrid traversal}                        & \multicolumn{1}{l|}{GNN $\rightarrow$ LSTM}                       & Y \\ \hline
\multicolumn{1}{|l|}{\multirow{2}{*}{LDC2017T10}}                                                                       & \multicolumn{1}{l|}{\citeb{bai:20} Graph reconstruction w/ node \& edge projection}           & \multicolumn{1}{l|}{T $\rightarrow$ T + reconstruction loss}      & Y \\ \cline{2-4} 
\multicolumn{1}{|l|}{}                                                                                                  & \multicolumn{1}{l|}{\citeb{mager:20} Fine-tuning GPT-2 on AMR-text joint distribution}          & \multicolumn{1}{l|}{GPT-2}                             & Y \\ \hline
\multicolumn{1}{|l|}{\multirow{9}{*}{WebNLG}}                                                                           & \multicolumn{1}{l|}{\citeb{marcheggiani:18} Graph encoding with GCNs}                                  & \multicolumn{1}{l|}{GCN $\rightarrow$ LSTM}                       & N \\ \cline{2-4} 
\multicolumn{1}{|l|}{}                                                                                                  & \multicolumn{1}{l|}{\citeb{distiawan:18} LSTM based triple encoder}                                 & \multicolumn{1}{l|}{LSTM $\rightarrow$ LSTM}                      & Y \\ \cline{2-4} 
\multicolumn{1}{|l|}{}                                                                                                  & \multicolumn{1}{l|}{\citeb{ferreira:19} Discrete neural pipelines \& comparisons to end-to-end}    & \multicolumn{1}{l|}{GRU $\rightarrow$ GRU + T}                    & Y \\ \cline{2-4} 
\multicolumn{1}{|l|}{}                                                                                                  & \multicolumn{1}{l|}{\citeb{moryossef:19} Sentence planning with ordered trees}                      & \multicolumn{1}{l|}{LSTM $\rightarrow$ LSTM}                      & Y \\ \cline{2-4} 
\multicolumn{1}{|l|}{}                                                                                                  & \multicolumn{1}{l|}{\citeb{ribeiro_a:20} Complementary graph contextualization}                     & \multicolumn{1}{l|}{GAT $\rightarrow$ T}                          & Y \\ \cline{2-4} 
\multicolumn{1}{|l|}{}                                                                                                  & \multicolumn{1}{l|}{\citeb{song:20} Detachable multi-view reconstruction}                      & \multicolumn{1}{l|}{T $\rightarrow$ T}                            & N \\ \cline{2-4} 
\multicolumn{1}{|l|}{}                                                                                                  & \multicolumn{1}{l|}{\citeb{zhao:20} Dual encoder for structure and planning}                   & \multicolumn{1}{l|}{GCN $\rightarrow$ LSTM}                       & Y \\ \cline{2-4} 
\multicolumn{1}{|l|}{}                                                                                                  & \multicolumn{1}{l|}{\citeb{ribeiro:21} Task-adaptive pretraining for PLMs}                        & \multicolumn{1}{l|}{BART + T5}                         & Y \\ \cline{2-4} 
\multicolumn{1}{|l|}{}                                                                                                  & \multicolumn{1}{l|}{\citeb{agarwal:21} Knowledge enhanced language models \& KeLM dataset}        & \multicolumn{1}{l|}{T5}                                & Y \\ 
\cline{2-4} 
\multicolumn{1}{|l|}{}                                                                                                  & \multicolumn{1}{l|}{\citeb{ke:21} Graph-text joint representations \& pretraining strategies}        & \multicolumn{1}{l|}{BART + T5}                                & Y \\\hline
\multicolumn{4}{|c|}{\textbf{Record-to-Text (Table-to-text)}}                                                                                                                                                                                                                         \\ \hline
\multicolumn{1}{|l|}{\multirow{17}{*}{WikiBio}}                                                                         & \multicolumn{1}{l|}{\citeb{lebret:16} Tabular positional embeddings \& WikiBio dataset}          & \multicolumn{1}{l|}{LSTM $\rightarrow$ LSTM + Kneser-Ney}         & N \\ \cline{2-4} 
\multicolumn{1}{|l|}{}                                                                                                  & \multicolumn{1}{l|}{\citeb{bao:18} Encoding tabular attributes \& WikiTableText dataset}      & \multicolumn{1}{l|}{GRU $\rightarrow$ GRU}                        & N \\ \cline{2-4} 
\multicolumn{1}{|l|}{}                                                                                                  & \multicolumn{1}{l|}{\citeb{liu:18} Field information through modified LSTM gating}            & \multicolumn{1}{l|}{LSTM $\rightarrow$ LSTM}                      & N \\ \cline{2-4} 
\multicolumn{1}{|l|}{}                                                                                                  & \multicolumn{1}{l|}{\citeb{sha:18} Link-based and content-based attention}                    & \multicolumn{1}{l|}{LSTM $\rightarrow$ LSTM}                      & N \\ \cline{2-4} 
\multicolumn{1}{|l|}{}                                                                                                  & \multicolumn{1}{l|}{\citeb{perez:18} Multi-instance learning w/ alignment-based rewards}        & \multicolumn{1}{l|}{LSTM $\rightarrow$ LSTM}                      & Y \\ \cline{2-4} 
\multicolumn{1}{|l|}{}                                                                                                  & \multicolumn{1}{l|}{\citeb{ma:19} Key fact identification and data augment for few shot}     & \multicolumn{1}{l|}{LSTM + T}                          & N \\ \cline{2-4} 
\multicolumn{1}{|l|}{}                                                                                                  & \multicolumn{1}{l|}{\citeb{liuhierarchical:19} Hierarchical encoding w/ supervised auxiliary learning}    & \multicolumn{1}{l|}{LSTM $\rightarrow$ LSTM}                      & Y \\ \cline{2-4} 
\multicolumn{1}{|l|}{}                                                                                                  & \multicolumn{1}{l|}{\citeb{liu_b:19} Forced attention for omission control}                     & \multicolumn{1}{l|}{LSTM $\rightarrow$ LSTM}                      & Y \\ \cline{2-4} 
\multicolumn{1}{|l|}{}                                                                                                  & \multicolumn{1}{l|}{\citeb{chen:19} External contextual information w/ knowledge graphs}       & \multicolumn{1}{l|}{GRU $\rightarrow$ GRU}                        & Y \\ \cline{2-4} 
\multicolumn{1}{|l|}{}                                                                                                  & \multicolumn{1}{l|}{\citeb{tian:19} Confidence priors for hallucination control}               & \multicolumn{1}{l|}{BERT + Pointer Networks}           & Y \\ \cline{2-4} 
\multicolumn{1}{|l|}{}                                                                                                  & \multicolumn{1}{l|}{\citeb{chen_b:20} Soft copy switching policy for few-shot learning}          & \multicolumn{1}{l|}{GPT-2}                             & Y \\ \cline{2-4} 
\multicolumn{1}{|l|}{}                                                                                                  & \multicolumn{1}{l|}{\citeb{ye:20} Variational autoencoders for template induction}           & \multicolumn{1}{l|}{VAE modified to VTM}               & Y \\ \cline{2-4} 
\multicolumn{1}{|l|}{}                                                                                                  & \multicolumn{1}{l|}{\citeb{wang:21} Autoregressive modeling with iterative text-editing}       & \multicolumn{1}{l|}{Pointer networks + Text editing}   & Y \\ \cline{2-4} 
\multicolumn{1}{|l|}{}                                                                                                  & \multicolumn{1}{l|}{\citeb{zhao:21} Reinforcement learning with adversarial networks}          & \multicolumn{1}{l|}{GAN}                               & N \\ \cline{2-4} 
\multicolumn{1}{|l|}{}                                                                                                  & \multicolumn{1}{l|}{\citeb{ghosh:21} Linearly combined multi-reward policy}                     & \multicolumn{1}{l|}{Pointer networks}                  & N \\ \cline{2-4} 
\multicolumn{1}{|l|}{}                                                                                                  & \multicolumn{1}{l|}{\citeb{yang:21} Source-target disagreement auxiliary loss}                 & \multicolumn{1}{l|}{T}                                 & N \\ \cline{2-4} 
\multicolumn{1}{|l|}{}                                                                                                  & \multicolumn{1}{l|}{\citeb{su:21} BERT-based IR system for contextual examples}              & \multicolumn{1}{l|}{T5 + BERT}                         & Y \\ \hline
\multicolumn{1}{|l|}{\multirow{9}{*}{Rotowire}}                                                                         & \multicolumn{1}{l|}{\citeb{wiseman:17} Classification-based metrics \& RotoWire dataset}          & \multicolumn{1}{l|}{LSTM $\rightarrow$ LSTM + Templated}          & N \\ \cline{2-4} 
\multicolumn{1}{|l|}{}                                                                                                  & \multicolumn{1}{l|}{\citeb{nie:18} Numeric operations and operation-result encoding}          & \multicolumn{1}{l|}{GRU $\rightarrow$ GRU + operation encoders}   & Y \\ \cline{2-4} 
\multicolumn{1}{|l|}{}                                                                                                  & \multicolumn{1}{l|}{\citeb{puduppully:19a} Dynamic hierarchical entity-modeling \& MLB dataset}       & \multicolumn{1}{l|}{LSTM $\rightarrow$ LSTM}                      & Y \\ \cline{2-4} 
\multicolumn{1}{|l|}{}                                                                                                  & \multicolumn{1}{l|}{\citeb{puduppully:19b} Content selection \& planning w/ gating and IE}            & \multicolumn{1}{l|}{LSTM $\rightarrow$ LSTM}                      & Y \\ \cline{2-4} 
\multicolumn{1}{|l|}{}                                                                                                  & \multicolumn{1}{l|}{\citeb{gong_b:20} Contextualized numeric representations}                    & \multicolumn{1}{l|}{LSTM $\rightarrow$ LSTM}                      & Y \\ \cline{2-4} 
\multicolumn{1}{|l|}{}                                                                                                  & \multicolumn{1}{l|}{\citeb{iso:20} Dynamic salient record tracking w/ stylized generation}    & \multicolumn{1}{l|}{GRU $\rightarrow$ GRU}                        & N \\ \cline{2-4} 
\multicolumn{1}{|l|}{}                                                                                                  & \multicolumn{1}{l|}{\citeb{rebuffel:20} Two-tier hierarchical input encoding}                      & \multicolumn{1}{l|}{T $\rightarrow$ LSTM}                         & N \\ \cline{2-4} 
\multicolumn{1}{|l|}{}                                                                                                  & \multicolumn{1}{l|}{\citeb{li:21} Auxiliary supervision w/ reasoning over entity graphs}     & \multicolumn{1}{l|}{LSTM + GAT}                        & Y \\ \cline{2-4} 
\multicolumn{1}{|l|}{}                                                                                                  & \multicolumn{1}{l|}{\citeb{puduppully:21} Paragraph-centric macro planning}                          & \multicolumn{1}{l|}{LSTM $\rightarrow$ LSTM}                      & Y \\ \cline{2-4} 
\multicolumn{1}{|l|}{}                                                                                                  & \multicolumn{1}{l|}{\citeb{puduppully:22} Interweaved plan and generation w/  variational models}                          & \multicolumn{1}{l|}{LSTM $\rightarrow$ LSTM}                      & Y \\ \hline
\end{tabular}
\end{table}
\begin{table}[htp]
\scriptsize
\centering
\begin{tabular}{|llll|}
\hline
\multicolumn{1}{|c|}{Dataset}                                                                                           & \multicolumn{1}{c|}{Publication Highlights}                                    & \multicolumn{2}{c|}{Framework \& Human Evaluation}         \\ \hline
\multicolumn{4}{|c|}{\textbf{Record-to-Text (Continued)}}                                                                                                                                                                                                                         \\ \hline
\multicolumn{1}{|l|}{TabFact}                                                                                           & \multicolumn{1}{l|}{\citeb{chen_c:20} Coarse-to-fine two-stage generation}                       & \multicolumn{1}{l|}{LSTM + T + GPT-2 + BERT}           & Y \\ \hline
\multicolumn{1}{|l|}{WikiPerson}                                                                                        & \multicolumn{1}{l|}{\citeb{wang:20} Disagreement loss w/ optimal-transport matching loss}      & \multicolumn{1}{l|}{T}                                 & Y \\ \hline
\multicolumn{1}{|l|}{Humans, Books \& Songs}                                                                            & \multicolumn{1}{l|}{\citeb{gong:20} Attribute prediction-based reconstruction loss}            & \multicolumn{1}{l|}{GPT-2}                             & Y \\ \hline
\multicolumn{1}{|l|}{\multirow{7}{*}{\begin{tabular}[c]{@{}l@{}}ToTTo\\ LogicNLG\\ NumericNLG\end{tabular}}}                                                                                     & \multicolumn{1}{l|}{\citeb{liu:21} Contextual examples through k nearest neighbors}           & \multicolumn{1}{l|}{GPT-3}                             & Y \\ \cline{2-4}
\multicolumn{1}{|c|}{}  & \multicolumn{1}{l|}{\citeb{suadaa:21} Targeted table cell representation}                        & \multicolumn{1}{l|}{GPT-2}                             & Y \\ \cline{2-4}
\multicolumn{1}{|c|}{}                                                                                       & \multicolumn{1}{l|}{\citeb{chen:21} Semantic confounders w/ Pearl's do-calculus}               & \multicolumn{1}{l|}{DCVED + GPT}                       & Y \\ \cline{2-4}
\multicolumn{1}{|c|}{}                                                                                                  & \multicolumn{1}{l|}{\citeb{plog:2022} Table-to-logic pretraining for logic text generation}              & \multicolumn{1}{l|}{T5 + BART}                         & Y \\ \cline{2-4}
\multicolumn{1}{|c|}{}                                                                                         & \multicolumn{1}{l|}{\citeb{r2d2:2023} Faithful generation with unlikelihood \& replacement detection}               & \multicolumn{1}{l|}{T5}                       & Y \\ \cline{2-4}
\multicolumn{1}{|c|}{}                                                                                         & \multicolumn{1}{l|}{\citeb{ttt_1:2022} Table serialization and structural encoding}               & \multicolumn{1}{l|}{T $\rightarrow$ GPT-2}                       & Y \\ \cline{2-4}
\multicolumn{1}{|c|}{}                                                                                         & \multicolumn{1}{l|}{\citeb{ttt_2:2022} T5 infused with tabular embeddings}               & \multicolumn{1}{l|}{T5}                       & N \\ \hline
\multicolumn{4}{|c|}{\textbf{Cross-domain}}                                                                                                                                                                                                                           \\ \hline
\multicolumn{1}{|l|}{\multirow{10}{*}{\begin{tabular}[c]{@{}l@{}}E2E\\ WebNLG\\ DART\\ WikiBio\\ RotoWire\\ WITA\end{tabular}}} & \multicolumn{1}{l|}{\citeb{jagfeld:18} Char-based vs word-based seq2seq}                          & \multicolumn{1}{l|}{GRU $\rightarrow$ GRU}                        & Y \\ \cline{2-4} 
\multicolumn{1}{|l|}{}                                                                                                  & \multicolumn{1}{l|}{\citeb{wiseman:18} Template induction w/ neural HSMM decoder}                 & \multicolumn{1}{l|}{HSMM}                              & N \\ \cline{2-4} 
\multicolumn{1}{|l|}{}                                                                                                  & \multicolumn{1}{l|}{\citeb{fu_b:20} Training w/ partially aligned dataset}                     & \multicolumn{1}{l|}{T $\rightarrow$ T + supportiveness}           & Y \\ \cline{2-4} 
\multicolumn{1}{|l|}{}                                                                                                  & \multicolumn{1}{l|}{\citeb{kasner:20} Iteratively editing templated text}                        & \multicolumn{1}{l|}{GPT-2 + LaserTagger}               & N \\ \cline{2-4} 
\multicolumn{1}{|l|}{}                                                                                                  & \multicolumn{1}{l|}{\citeb{harkous:20} RoBERTa-based semantic fidelity classifier}                & \multicolumn{1}{l|}{GPT-2 + RoBERTa}                   & Y \\ \cline{2-4} 
\multicolumn{1}{|l|}{}                                                                                                  & \multicolumn{1}{l|}{\citeb{chen:20} Knowledge-grounded pre-training \& KGTEXT dataset}         & \multicolumn{1}{l|}{T $\rightarrow$ T + GAT}                      & N \\ \cline{2-4}
\multicolumn{1}{|l|}{}                                                                                                  & \multicolumn{1}{l|}{\citeb{lin:20} Hybrid attention-copy for stylistic imitation}         & \multicolumn{1}{l|}{LSTM + T}                      & Y \\ \cline{2-4}
\multicolumn{1}{|l|}{}                                                                                                  & \multicolumn{1}{l|}{\citeb{asdot} Disambiguation and stitching with PLMs}             & \multicolumn{1}{l|}{GPT3 + T5}                          & N \\ \cline{2-4}
\multicolumn{1}{|l|}{}                                                                                                  & \multicolumn{1}{l|}{\citeb{duong:2023} Unified learning of D2T and T2D}             & \multicolumn{1}{l|}{T5 + VAE}                          & N \\ \cline{2-4}
\multicolumn{1}{|l|}{}                                                                                                  & \multicolumn{1}{l|}{\citeb{jolly:2022} Search and learn in a few-shot setting}             & \multicolumn{1}{l|}{T5 + Search \& Learn}                          & Y \\ \hline
\multicolumn{4}{|c|}{\textbf{Timeseries-to-text}}                                                                                                                                                                                                                     \\ \hline
\multicolumn{1}{|l|}{WebNLG \& DART}                                                                                    & \multicolumn{1}{l|}{\citeb{sharma:21} Open-domain transfer learning for time-series narration}   & \multicolumn{1}{l|}{BART + T5 + Timeseries analysis}   & Y \\ \hline
\multicolumn{4}{|c|}{\textbf{Chart-to-text}}                                                                                                                                                                                                                          \\ \hline
\multicolumn{1}{|l|}{Chart2Text}                                                                                        & \multicolumn{1}{l|}{\citeb{obeid:20} Preprocessing w/ variable substitution}                    & \multicolumn{1}{l|}{T $\rightarrow$ T}                            & Y \\
\hline
\multicolumn{1}{|l|}{Chart-to-text}                                                                                        & \multicolumn{1}{l|}{\citeb{kanthara:22} Neural baselines for Chart-to-text dataset}                    & \multicolumn{1}{l|}{LSTM + T + BART + T5}                            & Y \\ \hline
\end{tabular}
\end{table}

\section{Innovations in the Seq2Seq Framework}
Seq2Seq models (see \textsection 3.3), serve as the basis for neural NLG \cite{sutskever:11, cho:14, vaswani:17}. As such, to compare the efficacy of neural architectures for long-form D2T, Wiseman \textit{et. al.} \cite{wiseman:17} compare the performance of various seq2seq models to their templated counterparts on the RotoWire dataset. Based on their observations, the conditional copy model \cite{gul:16} performs the best on both word-overlap and \textit{extractive metrics} (see \textsection 6.2.1) compared to the standard attention-based seq2seq model \cite{bahdanau:15} and its joint copy variant \cite{gu:16}. Similarly, in an evaluation of 62 seq2seq, data-driven, and templated systems for the E2E shared task, Du{\v{s}}ek \textit{et. al.} \cite{duvsek:18} note that seq2seq systems dominate in terms of both automated word-based metrics and \textit{naturalness} in human judgement. Wiseman \textit{et. al.} \cite{wiseman:17}, however, note that the traditional templated generation models outperform seq2seq models on extractive metrics although they score poorly on word-overlap metirics. Thus, the adaptation of seq2seq models to D2T for richer narratives with less omissions and hallucinations still remains an active focus of the research community.

It is worth noting that all seq2seq models discussed below operate at the \textit{word} level. Models operating at the \textit{character} level \cite{goyal:16,agarwal:17, roberti:19} have shown reasonable efficacy with the added computational savings from forgoing the preprocessing steps of delexicalization and tokenization. However, the attention garnered by them from the research community is slim. From their comparative analysis, Jagfeld \textit{et. al.} \cite{jagfeld:18} note that as character-based models perform better on the E2E dataset while word-based models perform better on the more linguistically challenging WebNLG dataset, it is hard to draw conclusions on the framework most suited for generic D2T. In the sections that follow, we detail notable innovations over the last half-decade in seq2seq modeling, branched on the basis of their training strategies - \textit{supervised} and \textit{unsupervised} learning.

\subsection{Supervised Learning}

\subsubsection{\underline{Entity Encoders}}
\begin{figure}[h]
    \centering
    \includegraphics[scale=0.3]{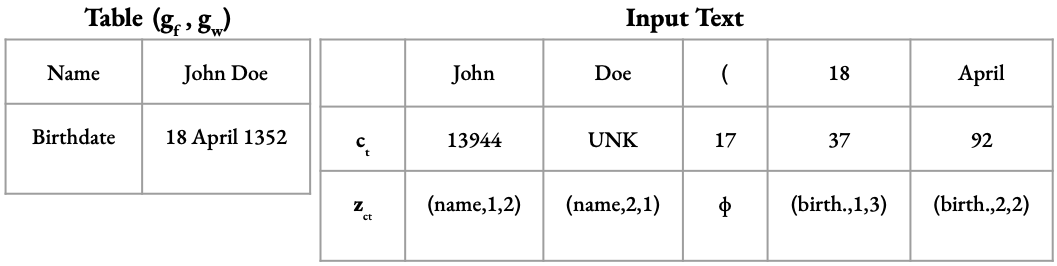}
    \caption{\footnotesize{Entity encoding scheme - Lebret \textit{et. al.} \cite{lebret:16}}}
    \label{fig:lebret}
\end{figure}

Centering theory \cite{grosz:95}, as well as many other noted linguistic frameworks \cite{ent:1,ent:2,ent:3,ent:4,ent:5,ent:6}, highlight the critical importance of \textit{entity mentions} to the coherence of the generated narrative. The ordering of these entities ($r.e$ in \textsection 3.3) is crucial for such narratives to be considered as \textit{entity coherent}  \cite{karamanis:04}. Unlike typical language models which are conditioned solely based on previously generated tokens $c_t$, Lebret \textit{et. al.} \cite{lebret:16} provide additional context $\{z_{c_t},g_f,g_w\}$ to the generation where $z_{c_t}$ represents table entity $c_t$ as a triplet of its corresponding field name, start, and end positions, and $\{g_f,g_w\}$ are one-hot encoded vectors where each element indicates the presence of table entities from the fixed field and word vocabularies - illustrated in Fig. \ref{fig:lebret}. Similarly, Bao \textit{et. al.} \cite{bao:18} encode the table cell $c$ and attributes $a$ as the concatenation $[e_{i}^{c}: e_{i}^{a}]$ where the decoder uses this vector to compute the attention weights. Liu \textit{et. al.} \cite{liu:18} modify the LSTM unit with a \textit{field} gate to update the cell memory indicating the amount of entity field information to be retained in the cell memory. Following \cite{lebret:16}, Ma \textit{et. al.} \cite{ma:19} use a Bi-LSTM to encode the concatenation of word, attribute and position embeddings. However, to indicate whether an entity is a \textit{key fact}, a multi-layer perceptron (MLP) classifier is used on said representation for binary classification. Inspired from Liu and Lapata \cite{liu_c:18}, Gong \textit{et. al.} \cite{gong:19} construct a historical timeline by sorting each table record with respect to its date field. Three encoders encode a table entity separately in row $r_{i,j}^{r}$, column $r_{i,j}^{c}$, and time $r_{i,j}^{t}$ dimensions. The concatenation of these representations are fed to a MLP to obtain a general representation $r_{i,j}^{gen}$ over which specialized attention weights are computed to obtain the final record representation as $\hat{r}_{i,j} = \alpha_{r}r_{i,j}^{r} + \alpha_{c}r_{i,j}^{c} + \alpha_{t}r_{i,j}^{t}$. Exploiting the attributes of the E2E dataset - the set number of unique MR attributes and the limited diversity in lexical instantiations of their values, Puzikov and Gurevych \cite{puzikov:18} employ a simple approach wherein the recurrent encoder is replaced with one dense layer that takes in MR representations through embedding lookup. Similarly, to keep track of entity mentions in the SF dataset for long-form text generation, Kiddon \textit{et. al.} \cite{kiddon:16} introduce a \textit{checklist} vector $a_{t}$ that aids two additional encoders to track used (mentioned in the resulting narrative) and new (not mentioned as of time step $t$) items on the defined agenda. The output hidden state is modeled as a linear interpolation between the three encoder states - $c_{t}^{gru}$ of the base GRU and $\{c_{t}^{new},c_{t}^{used}\}$ from the agenda models, weighted by a probabilistic classifier. Extending this concept of entity encoding to transformer-based architectures, Chen \textit{et. al.} \cite{ttt_1:2022} adapt the multi-headed attention layer architecture \cite{vaswani:17} to encode serialized table attributes that is then fed to a GPT-based decoder. Similarly, as a unified text-to-text alternative approach to \cite{ttt_1:2022}, Andrejczuk \textit{et. al.} \cite{ttt_2:2022} include the row and column embeddings $\hat{r}_{i,j}$ of the input table on top of the token embeddings for table-structure learning in a T5 model.

Certain D2T tasks, such as sport commentaries \cite{robin:95, tanaka:98, barzilay:05}, require reasoning over \textit{numeric} entities present in the input data instance. Although numeracy in language modeling is a prominent niche of its own \cite{num:1, num:2, num:3}, notable D2T-specific approaches include that of Nie \textit{et. al.} \cite{nie:18} - precomputing the results of numeric operations $op_{i} \in \{minus, argmax\}$ on the RotoWire dataset, the authors propose the combination of dedicated \textit{operation} and \textit{operation-result} encoders, the latter utilizing a quantization layer for mapping lexical choices to data values, in addition to a record encoder. In similar fashion to \cite{yang:17}, the concatenated embeddings $\{r.idx, r.e, r.m\}$ fed to a bi-directional GRU generate record representations while the concatenated embeddings of $op_{i}$ attributes fed to a non-linear layer yields operation representations. To address the difficulty in establishing lexical choices on sparse numeric values \cite{reiter:05, smiley:16}, the authors add quantization to the operation-results encoder that maps results of scalar operations $e$ (minus) to $l \in L$ possible bins through a weighed representation ($h_{i} = \sum_{l}\mu_{i,l} \ e$) using softmax scores of each individual result $\mu_{i,l}$. Following this body of work, to contextualize numeric representations and thus understand their logical relationships, Gong \textit{et. al.} \cite{gong_b:20} feed raw numeric embeddings for all numericals corresponding to the same table attributes to a transformer-based encoder to obtain their \textit{contextualized} representations. Through a ranking scheme based on a fully connected layer, these contextualized representations are further trained to favor larger numbers.

\subsubsection{\underline{Hierarchical Encoders}}
\begin{figure}[h]
    \centering
    \subfloat[\scriptsize \centering Plan Encoding]{{\includegraphics[scale=0.35]{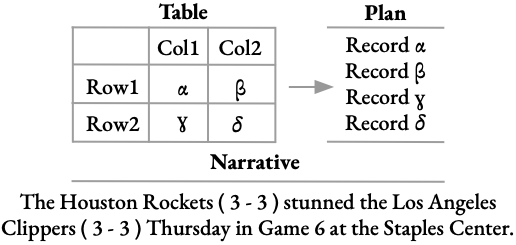} }}%
    \subfloat[\scriptsize \centering Hierarchical Encoding]{{\includegraphics[scale=0.3]{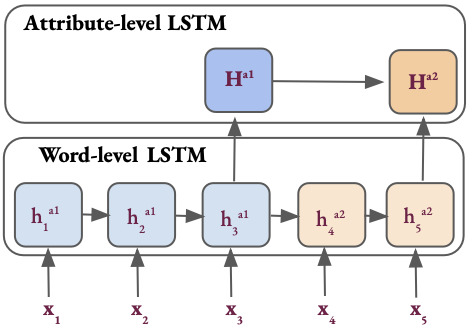} }}%
    \caption{\footnotesize Illustrations for plan \cite{puduppully:19b} and hierarchical \cite{liuhierarchical:19} encoding schemes.}%
    \label{fig:hiplans}
\end{figure}

The intuition behind the use of hierarchical encoders, in the context of D2T, is to model input representations at different granularities, either through dedicated modules \cite{zhang:18, rebuffel:20, liuhierarchical:19} or attention schemes \cite{yang:17,liu:18,puduppully:19a}. As such, Zhang \textit{et. al.} \cite{zhang:18} leverage their CAEncoder \cite{zhang:17} to incorporate precomputed future representations $h_{i+1}$ into current representation $h_{i}$ through a two-level hierarchy. Similarly, Rebuffel \textit{et. al.} \cite{rebuffel:20} propose a two-tier encoder to preserve the data structure hierarchy - the first tier encodes each entity $e_{i}$ based on its associated record embeddings $r_{i,j}$ while the second tier encodes the data structure based on its entity representation $h_{i}$ obtained through the individual embeddings $r_{i,j}$. On the other hand, Liu \textit{et. al.} \cite{liuhierarchical:19}, as illustrated in Fig. \ref{fig:hiplans}b, propose a word-level $h_{r.e}^{r.m}$ and an attribute-level $H^{r.e}$ two-encoder setup to capture the attribute-value hierarchical structure in tables. The attribute-level encoder takes in the last hidden state $h_{last}^{r.e}$ for attribute $r.e$ from the word level LSTM as its input. Using these hierarchical representations, fine-grained attention $\beta_{r.e}^{r.m} \propto g(h_{r.e}^{r.m}, s_t)$ and coarse-grained attention $\gamma^{r.e} \propto g(H^{r.e}, s_t)$ are used for decoding where $g$ represents a softmax function. Similarly, based on hierarchical attention \cite{yang:17}, Liu \textit{et. al.} \cite{liu:18} employ an attention scheme that attends to both word level and field level tokens. Following this, Puduppully \textit{et. al.} \cite{puduppully:19a} propose language modeling conditioned on both the data instance and a dynamically updated entity representation. At each time-step $t$, a gate $\gamma_{t}$ is used to decide whether an update is necessary for the entity memory representation $u_{k}$ and a parameter $\delta_{t,k}$ decides the impact of said update (\ref{eqn:ent1}).
\begin{equation}
\label{eqn:ent1}
    \gamma_{t} = \sigma(W_{1}s_{t} + b_{1}) \ \& \
    \delta_{t,k} = \gamma_{t} \odot \sigma(W_{2}s_{t} + W_{3}u_{t-1,k} + b_{3})    
\end{equation}

\subsubsection{\underline{Plan Encoders \& Autoencoders}}
Traditionally, the \textit{what to say } aspect of D2T (see \textsection 3.1) used to be its own module in a set of pipelines \cite{reiterdale:97, survey:18}, thus offering flexibility in planning the narrative structure. However, the end-to-end learning paradigm often models content selection and surface realization as a shared task \cite{lebret:16, mei:16, wiseman:17}. Although convenient, without \textit{explicitly} modeling the planning of the narrative (Fig. \ref{fig:hiplans}a), language models struggle to keep coherence in long-form generation tasks. As such, Puduppully \textit{et. al.} \cite{puduppully:19b} model the generation probability $P(y|r)$ as the joint probability of narrative $y$ and content plan $z$ given a record $r$ such that $P(y|r) = \sum_{z} P(z|r)P(y|r,z)$. Similar to their prior work \cite{puduppully:19a}, a content selection gate operates over the record representation $r_{i}$ giving an information controlled representation $r_{i}^{cs}$. The elements of $z$ are extracted using an information extraction system \cite{wiseman:17} and correspond to entities in $y$ while pointer networks \cite{vinyals:15} are use to align elements in $z$ to $r$ during training. Iso \textit{et. al.} \cite{iso:20}, on the other hand, avoid precomputing content plans $z$ by dynamically choosing data records during decoding - an additional memory state $h^{ent}$ remembers mentioned entities and updates the language model state $h^{lm}$ accordingly. The authors propose two representations for an entity  - static embedding $e$ based on row $r_{i}$ and aggregated embedding $\bar{e}$ based on all rows where the entity appears. In the context of the RotoWire dataset, the aggregate embedding $\bar{e}$ is supposed to represent how entity $e$ played in the game. For $h_{t} = \{h_{t}^{lm},h_{t}^{ent}\}$, $P(z_{t}=1|h_{t-1})$ (\ref{iso:1}) models the transition probability, and based on whether $e$ belongs to the set of entities $\epsilon_{t}$ that have already appeared at time step $t$, $P(e_{t} = e | h_{t-1})$ (\ref{iso:2}) computes the next probable entity $e$ to mention. The authors note that such discrete tracking dramatically suppresses the generation of redundant relations in the narrative. 
\begin{align}
    \label{iso:1}
    P(z_{t}=1|h_{t-1}) &= \sigma(W_{1}(h_{t-1}^{lm} \oplus h_{t-1}^{ent})) \\
    \label{iso:2}
    P(e_{t} = e | h_{t-1}) &\propto
    \begin{cases}
        e^{(h_{s}^{ent}W_{1}h_{t-1}^{lm})} \ e \in \epsilon_{t-1} \\
        e^{(\bar{e}W_{2}h_{t-1}^{lm})} \ otherwise
    \end{cases}
\end{align}
With the premise that paragraphs are the smallest sub-categorization where coherence and topic are defined \cite{zadrozny:91}, Puduppully and Lapata \cite{puduppully:21} propose a paragraph-based \textit{macro} planning framework specific to the design of MLB \cite{puduppully:19a} and RotoWire \cite{wiseman:17} datasets where the input to the seq2seq framework are predicted macro-plans (sequence of paragraphs). Building upon this, in contrast to precomputing global macro plans, Puduppully \textit{et. al.} \cite{puduppully:22} interweave the macro planning process with narrative generation where latent plans are sequentially inferred through a structured variational model as the narrative is generated conditioned on the plans so far and the previously generated paragraphs. Similarly, to establish order in the generation process, Sha \textit{et. al.} \cite{sha:18} incorporate link-based attention \cite{graves:16} in addition to content-based attention \cite{bahdanau:15} into their framework. Similar to transitions in Markov chains \cite{karlin:14}, a link matrix $\mathbb{L} \in \mathbb{R}^{n_{f} \times n_{f}}$ for $n_{f}$ tabular attributes defines the likelihood of transitioning from the mention of attribute $i$ to $j$ as $\mathbb{L}(f_{j},f_{i})$. Wang \textit{et. al.} \cite{wang:21} propose combining autoregressive modeling \cite{see:17} to generate skeletal plans and using an iterative text-editing based non-autoregressive decoder \cite{gu:19} to generate narratives constrained on said skeletal plans. The authors note that this approach reduces hallucination tendencies of the model. Similarly, motivated by the strong correlation observed between entity-centric metrics for record coverage and hallucinations, Liu \textit{et. al.} \cite{liu_b:21} adopt a two-stage generation process where a plan generator first transforms the input table records into serialized plans $R \rightarrow R + P$ based on the separator token $SEP$ and then translates the plans into narratives with the help of appended auxiliary entity information extracted through NER. 

\textit{Handcrafted templates} traditionally served as pre-defined structures where entities computed through content selection would be plugged-in. However, even in the neural D2T paradigm, inducing underlying templates helps capture the narrator voicing and stylistic representations present in the training set. As such, Ye \textit{et. al.} \cite{ye:20} extend the use of the variational autoencoders (VAEs) \cite{kingma:14} for template induction with their variational template machine (VMT) that disentangles the latent representation of the template $z$ and the content $c$. In essence, the model can be trained to follow specific templates by sampling from $z$. Inspired from stylistic encoders \cite{hu:17}, the authors further promote template learning by anonymizing entities in the input table thus effectively masking the content selection process. Similarly, to mitigate the strong model biases in the standard conditional VAEs \cite{tomczak:18}, Chen \textit{et. al.} \cite{chen:21} estimate semantic \textit{confounders} $z_c$ - linguistically similar entities to the target tokens that confound the logic of the narrative. Compared to the standard formulation $p(y|x)$, the authors employ Pearl's do-calculus \cite{pearl:10} to learn the objective $p(y|\mathrm{do}(x))$ that asserts that confounder $z_c$ is no longer determined by instance $x$, thus ensuring logical consistency in the narrative. To ensure that the estimated confounders are meaningful, they are grounded through proxy variables $c$ such that confounding generation $p(c|z_{m})$ can be minimized. Recently, modeling D2T and T2D as complementary tasks, Doung \textit{et. al.} \cite{duong:2023} leverage the VAE formulation with the underlying architecture of a pre-trained T5 model to offer a unified multi-domain framework for the dual task. To combat the lack of parallel-corpora for the back-translation (T2D) training, the authors introduce latent variables to model the marginal probabilities of back-translation through an iterative learning process. 

Likewise, for approaches beyond the use of autoencoders, Chen \textit{et. al.} \cite{chen_c:20} take inspiration from practices in semantic parsing \cite{dong:18} and propose a coarse-to-fine two-stage generation scheme. In the first stage, a template $Y_{T}$ containing placeholder tokens $ENT$ is generated, representing the global logical structure of the narrative. The entities are then copied over from the input data instance to replace tokens $ENT$ in the second step to generate the final narrative $\hat{Y}$. Suadaa \textit{et. al.} \cite{suadaa:21}, similarly, follow template-guided generation \cite{kale:20} (see \textsection 4.2) where the precomputed results of numeric operations are copied over to the template and replace the placeholder tokens. For Pre-trained Language Models (PLMs), the authors incorporate copying into the fine-tuning stage for this action. 

\subsubsection{\underline{Stylistic Encoders}}
\begin{figure}[h]
    \centering
    \includegraphics[scale=0.45]{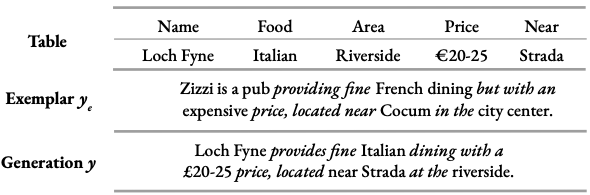}
    \caption{\footnotesize{Style imitation with exemplar narratives - Lin \textit{et. al.} \cite{lin:20}}}
    \label{fig:lin}
\end{figure}

In addition to the traits of coherence, fluency, and fidelity, stylistic variation is crucial to NLG \cite{stent:05}. It is interesting to note that the n-gram entropy of generated texts in seq2seq based NLG systems are significantly lower than that in its training data - leading to the conclusion that these systems adhere to only a handful of dominant patterns observed in the training set \cite{oraby:18}. Thus, introducing control measures to text generation has recently garnered significant attention from the NLG community \cite{hu:17,weng:21}. As such, the semantically conditioned LSTM (SC-LSTM) proposed by Wen \textit{et. al.} \cite{wen:15} extends the LSTM cell to incorporate a one-hot encoded MR vector $d$ that takes the form of a sentence planner. Following this, Deriu and Cieliebak \cite{deriu:18} append additional syntactic control measures to the MR vector $d$ (such as the first token to appear in the utterances and expressions for different entity-value pairs) by simply appending one-hot vectors representation of these control mechanisms to $d$. Similarly, Lin \textit{et. al.} \cite{lin:20} tackle the lack of a template-based parallel dataset with \textit{style imitation} - as illustrated in Fig. \ref{fig:lin}, for each instance $(x,y)$, an \textit{exemplar narrative} $y_{e}$ is retrieved from the training set based on field-overlap distance $D(x,x_{e})$ and an additional encoder is used to encode $y_{e}$. The model is trained with competing objectives for content determination $P(y|x,y_{e})$ and style embodiment $P(y_{e}|x_{e},y_{e})$ with an additional content coverage constraint for better generation fidelity.

\subsubsection{\underline{Graph Encoders}}
The use of explicit graph encoders in D2T stems from the intuition that neural graph encoders such as Graph Convolutional Networks (GCNs) \cite{kipf:16} have strong relational inductive baises that produce better representations of input graphs \cite{battaglia:18} as an effective alternative to linearization. This entails generating representations for the nodes $v \in V$ and edges $(u,v) \in E$ in the input graph. 

\vspace{0.25cm}
\par\noindent \textbf{GCNs \& Graph-RNNs:} Marcheggiani and Titov \cite{marcheggiani:17} compute node represenations $h_{v}^{'}$ (\ref{gcn_1}) through explicit modeling of edge labels $lab(u,v)$ and directions $dir(u,v) \in \{in, out, loop \}$ for each neighboring node $u \in N(v)$ in their GCN parameterization where learned scalar gates $g_{u,v}$ weigh the importance of each edge. With residual ($h_{v}^{r} = h_{v}^{'} + h_{v}$)  \cite{he:16} and dense ($h_{v}^{d} = [h_{v}^{'}; h_{v}]$) \cite{huang:17} skip connections, Marcheggiani and Perez-Beltrachini \cite{marcheggiani:18} adopt the above-mentioned encoder with an LSTM decoder \cite{luong:15} for graph-to-text generation. Differing from previous iterations of Graph LSTMs \cite{liang:16}, Distiawan \textit{et. al.} \cite{distiawan:18} compute the hidden states of graph entities with consideration of the edges pointing to the entity from the previous entities, allowing their GTR-LSTM framework to handle non-predefined relationships. The ordering of the vertices fed into the LSTM is based on a combination of topological sort and breath-first traversal. Inspired by hybrid traversal techniques \cite{shieber:90, narayan:12}, Ribeiro \textit{et. al.} \cite{ribeiro:19} propose a dual graph encoder - first operating on a top-down traversal of the input graph where the predicate $p$ between two nodes is used to transform labelled edges $(u_{i}, p, u_{j})$ to two unlabelled edges $(u_{i}, p)$ and $(p, u_{j})$ while the second operates on a bottom-up traversal where directions of edges are reversed $(u_{i}, u_{j}) \rightarrow (u_{j}, u_{i})$.
\begin{equation}
\label{gcn_1}
    h_{v}^{'} = \mathrm{ReLU}(\sum_{u \in N(v)}g_{u,v}(W_{dir(u,v)}h_{u} + b_{lab(u,v)}))    
\end{equation}
Damonte and Cohen \cite{damonte:19} note that GCNs can assist LSTMs in capturing re-entrant structures and long term dependencies. As such, to bridge the gap between the GCN \cite{beck:18, schlichtkrull:18} and the linearized LSTM encoders in graph-to-text translation, Zhao \textit{et. al.} \cite{zhao:20} propose DualEnc which uses both to capture their complementary effects. The first GCN models the graph, retaining its structural integrity, while the second GCN serializes and re-orders the graph nodes resembling a planning stage and feeds it to the LSTM decoder.

\vspace{0.2cm}
\par\noindent \textbf{GATs \& Graph Transformers:} To address the shortcomings of RNN-based sequential computing, Koncel-Kedziorski \textit{et. al.} \cite{koncel:19} extend the transformer architecture \cite{vaswani:17} to graph-structured inputs with the GraphWriter. The distinction of GraphWriter from graph attention networks (GAT) \cite{velivckovic:18} is made through the contextualization of each node representation $v_{i}$ (\ref{koncel}) with respect to its neighbours $u_{j} \in N(v_{i})$ through attention mechanism $a_{n}$ for the $\mathcal{N}$ attention heads. In contrast, Ribeiro \textit{et. al.} \cite{ribeiro_a:20} focus on capturing complementary graph contexts through distinct global $h_{v}^{global}$ and local $h_{v}^{local}$ message passing using GATs. Their approach to graph modeling also differs in its token-level approach for node representations with positional embeddings injected to preserve sequential order of the tokens.
\begin{equation}
\label{koncel}
    \hat{v_{i}} = v_{i} +  \Arrowvert_{n=1}^{\mathcal{N}}\sum_{N(v_{i})}\alpha_{ij}^{n}W_{v}^{n}u_{j} \quad \& \quad
    \alpha_{ij}^{n} = a^{n}(v_{i}, u_{j})
\end{equation}
Song \textit{et. al.} \cite{song:20} enrich the training signal to the relation-aware transformer model \cite{zhu:19} through additional multi-view autoencoding losses \cite{rei:17}. This detachable multi-view framework deconstructs the input graph into triple sets for the first view, reconstructed with a deep biaffine model \cite{dozat:17}, and linearizes the graph through a depth-first traversal for the second view. In contrast, Ke \textit{et. al.} \cite{ke:21} obtain the entity and relation embeddings through contextual semantic representations with their structure-aware semantic aggregation module added to each transformer layer - the module consists of a mean pooling layer for entity and relation representations, a structure-aware self-attention layer \cite{shaw:18}, and finally a residual layer that fuses the semantic and structural representations of entities.

\subsubsection{\underline{Reconstruction \& Hierarchical Decoders}}

\vspace{0.25cm}
\par\noindent \textbf{Input Reconstruction:} Conceptualized from autoencoders \cite{auto:1,auto:2,auto:3}, reconstruction-based models quantify the faithfulness of an encoded representation by correlating the decoded representation to the original input. As such, Wiseman \textit{et. al.} \cite{wiseman:17} adopt decoder reconstruction \cite{tu:17} to the D2T paradigm by segmenting the decoder hidden states $h_t$ into $\frac{T}{B}$ continuous blocks $b_i$ of size at most $B$. The prediction of record $r$ from such a block $b_i$, $p(r.e,r.m | b_{i})$, is modeled as softmax($f(b_{i})$) where $f$ is a convolutional layer followed by a multi-layer perceptron (MLP). To replicate the actions of an autoencoder, Chisholm \textit{et. al.} \cite{chisholm:17} train a seq2seq based reverse re-encoding \textit{text-to-data} model along with a forward seq2seq data-to-text model. Similarly, Roberti \textit{et. al.} \cite{roberti:19} propose a character-level GRU implementation where the recurrent module is passed as a parameter to either the encoder or the decoder depending on the forward $\hat{y} = f(x)$ or reverse $\hat{x} = g(y)$ direction. Following the mechanics of back-translation (text-to-data) \cite{sennrich:16, tu:17}, Bai \textit{et. al.} \cite{bai:20} extend the standard transformer decoder \cite{vaswani:17} to reconstruct the input graph by jointly predicting the node and edge labels while predicting the next token. The standard training objective of minimizing the negative log-likelihood of the conditional word probabilities $l_{std}$ is appended by a node prediction loss $l_{node}$ that minimizes the word-to-node attention distance and a edge prediction loss $l_{edge}$ that minimizes the negative log-likelihood over the projected edges. For table-structure reconstruction, Gong \textit{et. al.} \cite{gong:20} define the reconstruction loss based on attribute prediction and content matching similar to the optimal transport distance \cite{wang:20}. It should be noted that these auxiliary tasks improve the model performance in few-shot settings.

\vspace{0.25cm}
\par\noindent\textbf{Hierarchical Decoding:} Similar to hierarchical encoding (see \textsection 5.1.2), hierarchical \textit{decoding} intends to designate granular roles to each decoder in the hierarchy. Serban \textit{et. al.} \cite{serban:17} show that injecting variations at the conditional output distribution does not capture high-level variations. As such, to model both high and low level variations, Shao \textit{et. al.} \cite{shao:19} propose their planning-based hierarchical variational model (PHVM) based on the conditional variational auto-encoder \cite{sohn:15}. PHVM follows a hierarchical multi-step encoder-decoder setup where a plan decoder first generates a subset $g$ of the input $\{d_{i},...,d{n}\}\in x$. Then, in the hierarchical generation process, a \textit{sentence decoder} and a \textit{word decoder} generate the narrative conditioned on plan $g$. To dissipate the decoder responsibilities in the seq2seq paradigm, Su \textit{et. al.} \cite{su:18} propose a 4-layer hierarchical decoder where each layer is responsible for learning different parts of the output speech. The training instances are appended with part-of-speech (POS) tags such that each layer in the decoder hierarchy is responsible for decoding words associated with a specific set of POS patterns.

\vspace{0.25cm}
\par\noindent\textbf{Hierarchical Attention-based Decoding:} To alleviate omissions in narrative generation, Liu \textit{et. al.} \cite{liu_b:19} propose \textit{forced attention} - with word-level coverage $\theta_{t}^{i}$ and attribute-level coverage $\gamma_{t}^{e}$, a new context vector $\hat{c_{t}} = \pi c_{t} + (1-\pi)v_{t}$ is defined with a learnable vector $\pi$ and a compensation vector $v_{t} = f(\theta_{t}^{i},\gamma_{t}^{e})$ for low-coverage attributes $e$. To enforce this at a global scale, similar to Xu \textit{et. al.} \cite{xu:15}, a loss function $\mathbb{L}_{FA}$ based on $\gamma_{t}^{e}$ is appended to the seq2seq loss function.

\subsubsection{\underline{Regularization Techniques}}
Similar to regularization in the greater deep learning landscape \cite{reg}, regularization practices in D2T append additional constraints to the loss function to enhance generation fidelity. As such, Mei \textit{et. al.} \cite{mei:16} introduce a coarse-to-fine aligner to the seq-to-seq framework that uses a \textit{pre-selector} and \textit{refiner} to modulate the standard aligner \cite{bahdanau:15}. The pre-selector assigns each record a probability $p_{i}$ of being selected based on which the refiner re-weighs the standard aligner's likelihood $w_{ti}$ to $\alpha_{ti}$. The weighted average $z_{t} = \sum_{i}\alpha_{ti}m{i}$ is used as a soft approximation to maintain the architecture differentiability. Further, the authors regularize the model with a summation of the learned priors $\sum_{i=1}^{N}p_{i}$ as an approximation of the number of selected records. Similarly, Perez-Beltrachini and Lapata \cite{perez:18} precompute binary alignment labels for each token in the output sequence indicating its alignment with some attribute in the input record. The prediction of this binary variable is used as an auxiliary training objective for the D2T model. For tabular datasets, Liu \textit{et. al.} \cite{liuhierarchical:19} propose a two-level hierarchical encoder that breaks the learning of semantic tabular representation into three auxiliary tasks incorporated into the loss function of the model. The auxiliary sequence labeling task $L_{SL}$, learnt in unison with seq2seq learning, predicts the attribute name for each table cell. Similarly, the auto-encoder supervision $L_{AE}$ penalizes the distance between the table $z_t$ and the narrative $z_b$ representations, while the multi-label supervision task $L_{ML}$ predicts all the attributes in the given table. The individual losses, along with the language modeling loss, defines the loss function of the framework. To mitigate information hallucination and avoid the high variance exhibited by the use of policy gradients in the reinforcement-learning paradigm, Wang \textit{et. al.} \cite{wang:20} compute two losses in addition to the language modeling loss - the first checks the disagreement between the source table and the corresponding narrative through the L2 loss between their embeddings, similar to Yang \textit{et. al.} \cite{yang:21}, while the second uses optimal-transport \cite{chen:18} based maximum flow between the narrative and input distributions $\mu$ and $v$. Tian \textit{et. al.} \cite{tian:19} propose the use of \textit{confidence priors} to mitigate hallucination tendencies in table-to-text generation through learned confidence scores. At each decoding step $y_t$, instead of concatenating all the previous attention weights, only the antecedent attention weight $a_{t-1}$ is fed back to the RNN, such that an attention score $A_{t}$ can be used to compute how much $a_{t}$ affects the context vector $c_{t}$ - as all the source information in $c_{t}$ comes from $a_{t}$. The confidence score $C_t(y_{t})$ is then used to sample target sub-sequences faithful to the source using a variational Bayes scheme \cite{koller:09}. Similarly, inspired by Liu \textit{et. al.} \cite{liuhierarchical:19}, Li \textit{et. al.} \cite{li:21} propose two auxillary supervision tasks incorporated into the training loss - number ranking and importance ranking, both crucial to sport summaries, modeled with pointer networks on the outputs of the row and column encoders respectively.

\subsubsection{\underline{Reinforcement Learning}}
In the D2T premise, language-conditional reinforcement learning (RL) \cite{reinforcementsurvey} often aids in model optimization through its role as auxiliary loss functions. While traditionally, the BLEU (see \textsection 6.1) and TF-IDF \cite{tfidf} scores of generated texts were used as the basis for reinforcement learning \cite{liu_b:19}, Perez-Beltrachini and Lapata \cite{perez:18} use alignment scores of the generated text with the target text. Similarly, Gong \textit{et. al.} \cite{gong_b:20} use four entity-centric metrics that center around entity importance and mention. Rebuffel \textit{et. al.} \cite{rebuffel_b:20} propose a model agnostic RL framework, PARENTing which uses a combination of language model loss and RL loss computed based on PARENT F-score \cite{dhingra:19} to alleviate hallucinations and omissions in table-to-text generation. To avoid model overfitting on weaker training samples and to ensure the rewards reflect improvement made over pretraining, self-critical training protocol \cite{rennie:16} is applied using the REINFORCE algorithm \cite{williams:91}. The improvement in PARENT score over a randomly sampled candidate $y_{c}$ and a baseline sequence generated using greedy decoding $y_{b}$ is used as the reward policy. In contrast, Zhao \textit{et. al.} \cite{zhao:21} use generative adversarial networks (GANs) \cite{goodfellow:14} where the generator is modeled as a policy with the current state being the generated tokens and the action defined as the next token to select. The reward for the policy is a combination of two values - the discriminator probability of the sentence being real and the correspondence between generated narrative and the input table based on the BLEU score. As RL frameworks based on singular metrics makes it difficult to simultaneously tackle the multiple facets of generation, Ghosh \textit{et. al.} \cite{ghosh:21} linearly combine metrics for recall, repetition, and reconstruction, along with the BLEU score, to form a composite reward function. The policy is adapted from Wang \textit{et. al.} \cite{wang:18} and trained using Maximum Entropy Inverse Reinforcement Learning (MaxEnt IRL) \cite{ziebart:08}.

\subsubsection{\underline{Fine-tuning Pretrained Language Models}}
Pretrained lanuguage models (PLMs) \cite{devlin:19, radford:19} have been successful in numerous text generation tasks \cite{see:19,zhang_c:20}. The extensive pretraining grants these models certain worldly knowledge \cite{petroni:19} such that, at times, the models refuse to generate nonfactual narratives even when fed deliberately corrupted inputs \cite{ribeiro:21}. As such, Mager \textit{et. al.} \cite{mager:20} propose an alternate approach to fine-tuning GPT-2 for AMR-to-text generation where the fine-tuning is done on the joint distribution of the AMR $x_{j}$ and the text $y_{i}$ as $\prod_{i}^{N}p(y_{i}|y_{<i},x_{1:M}) \cdot \prod_{j}^{M}p(x_{j}|x_{<j})$. On the other hand, inspired by \textit{task-adaptive} pretraining strategies for text classification \cite{gururangan:20}, Ribeiro \textit{et. al.} \cite{ribeiro:21} introduce supervised and unsupervised task-adaptive pretraining stages as intermediaries between the original pretraining and the fine-tuning for graph-to-text translation. Interestingly, the authors note good performance of the task-adapted PLMs even when trained on shuffled graph representations. Chen \textit{et. al.} \cite{chen_b:20} note the few-shot learning capabilities of GPT-2 for table-to-text generation when appended with a soft switching policy for copying tokens \cite{see:17}. Similarly, as a light-weight alternative to fine-tuning the entire model, Li and Liang \cite{li_b:21} take inspiration from prompting \cite{brown:20}, and propose \textit{prefix-tuning} which freezes the model parameters to only optimize the prefix, a task-specific vector prepended to the input. The authors note significant improvements in low-data settings when the prefix is initialized with embeddings from task specific words such as \textit{table-to-text}. For avenues that allow leveraging the wordly knowledge of PLMS even \textit{without fine-tuning}, Xiang \textit{et. al.} \cite{asdot} leverage a combination of prompting GPT-3 for disambiguation and T5 for sentence-fusion leading to a domain-agnostic framework for data-to-text generation. 

Inspired from practices in unlikelihood learning \cite{unlikelihood:1, unlikelihood:2}, Nan \textit{et. al.} \cite{r2d2:2023} model T5 as both a generator and a faithfulness discriminator with two additional learning objectives for unlikelihood and replacement detection. To train the model with said objectives, $n$ contradictory sentences $Y^{(i,j)}_{False}$ are generated for each entailed sentence $Y^{(i)}_{True}$ wherein the discrimination probability is computed at every step of token generation. Similarly, to address omissions in D2T (\textsection 3.2), Jolly \textit{et. al.} \cite{jolly:2022} adapt the search-and-learn formulation \cite{li_search:2022} to a few-shot setting through a two-step finetuning process wherein a T5 model fine-tuned on the D2T task is further fine-tuned again with omitted attributes $r.e/r.m$ reinserted to the narratives as pseudo-grouthtruths.

\subsubsection{\underline{Supplemental Frameworks}}
\textbf{Supplementary Modules:} Fu \textit{et. al.} \cite{fu_b:20} propose the adaptation of the seq2seq framework for their partially-algined dataset WITA using a supportiveness adaptor and a rebalanced beam search. The pre-trained adaptor calculates supportiveness scores for each word in the generated text with respect to the input. This score is incorporated into the loss function of the seq2seq module and used to rebalance the probability distributions in the beam search. Framing the generation of narratives as a \textit{sentence fusion} \cite{barzilay_b:05} task, Kasner and Du{\v{s}}ek \cite{kasner:20} use the pre-trained LasterTagger text editor \cite{malmi:19} to iteratively improve a templated narrative. Su \textit{et. al.} \cite{su:21} adopt a BM25 \cite{robertson:09} based information retrieval (IR) system to their prototype-to-generate (P2G) framework, which, aided with their BERT-based prototype selector, retrieves contextual samples for the input data instance from Wikipedia, allowing for successful few shot learning in T5. For reasoning over tabulated sport summaries, Li \textit{et. al.} \cite{li:21} propose a variation on GATs named as GatedGAT that operates over an entity graph modeled after the source table to aid the generation model in entity-based reasoning.

\vspace{0.25cm}
\par\noindent\textbf{Re-ranking \& Pruning:} Du{\v{s}}ek and Jurcicek \cite{duvsek:16} append the seq2seq paradigm with an RNN-based re-ranker to penalize narratives with missing and/or irrelevant attributes from the beam search output. Based on the Hamming distance between two 1-hot vectors representing the presence of slot-value pairs, the classifier employs a logistic layer for a binary classification decision. Their framework, TGen, was the baseline for the E2E challenge \cite{duvsek:18}. Following this, Juraska \textit{et. al.} \cite{juraska:18} first compute slot-alignment scores with a heuristic-based slot aligner with is used to augment the probability score from the seq2seq model. The aligner consists of a \textit{gazetteer} that searches for overlapping content between the MR and its respective utterance, WordNet \cite{fellbaum:98} to account for semantic relationships, and hand-crafted rules to cover the outliers. Noting that even copy-based seq2seq models tend to omit values from the input data instance, Gehrmann \textit{et. al.} \cite{gehrmann:18} incorporate coverage $cp$ (\ref{eq:9}) and length $lp$ (\ref{eq:10}) penalties of Wu \textit{et. al.} \cite{gnmt:16}. With tunable parameters $\alpha$ and $\beta$, $cp$ increases when too many generated words attend to the same input $a_{i}^{t}$ and $lp$ increases with the length of the generated text. In contrast to Tu \textit{et. al.}\cite{tu:16}, however, the penalties are only used during inference to re-rank the beams. Similar to Paulus \textit{et. al.} \cite{paulus:18}, authors prune beams that start with the same bi-gram to promote syntactic variations in the generated text. Similar to NLI based approaches, Harkous \textit{et. al.} \cite{harkous:20} append a RoBERTa \cite{liu:19} based semantic fidelity classifier (SFC) that reranks the beam output from a fine-tuned GPT-2 model. 
\begin{align}
    \label{eq:9}
    cp(x,y) &= \beta \cdot \sum_{i=1}^{|x|} \mathrm{log}(\mathrm{min}(\sum_{t=1}^{|y|}a_{i}^{t},1)) \\
    \label{eq:10}
    lp(y) &= \frac{(5+|y|)^{\alpha}}{(5+1)^{\alpha}}
\end{align}

\subsubsection{\underline{Ensemble Learning}}
Juraska \textit{et. al.} \cite{juraska:18} propose SLUG, an ensemble of three neural encoders - two LSTMs and one CNN \cite{lecun:98}, individually trained, for MR-text generation. The authors note that selecting tokens with the maximum log-probability by averaging over different encoders at each time steps results in incoherent narratives, thus the output candidate is selected based on the rankings among the top 10 candidates from each model. SLUG's ensemble along with its data preprocessing and ranking schemes, as detailed in the sections above, was crowned winner of the 2017 E2E challenge \cite{duvsek:18}. Similarly, to prompt the models $f_{1},...,f_{n}$ in the ensemble to learn distinct sentence templates, Gehrmann \textit{et. al.} \cite{gehrmann:18} adopt diverse ensembling \cite{guzman:12} where an unobserved random variable $w\sim\mathrm{Cat}(1/n)$ assigns a weight to each model for each input. Constraining $w$ to $\{0,1\}$ trains each model $f_{i}$ on a subset of the training set thus leading to each model learning distinct templates. The final narrative is generated with a single model $f$ with the best perplexity on the validation set.

\subsection{Unsupervised Learning}

\subsubsection{\underline{D2T Specific Pretraining}}
Following the successful applications of knowledge-grounded language models \cite{ahn:16, logan:19}, Konstas \textit{et. al.} \cite{konstas:17} propose a domain-specific pretraining strategy inspired by Sennrich \textit{et. al.} \cite{sennrich:16} to combat the challenges in data sparsity, wherein self-training is used to bootstrap an AMR parser from the large unlabeled Gigaword corpus \cite{napoles:12} which is in turn used to pretrain an AMR generator. Both the generator and parser adopt the stacked-LSTM architecture \cite{gnmt:16} with a global attention decoder \cite{luong:15}. Similarly, following success brought forth by the suite of \textit{pre-trained language models} (PLMs) \cite{devlin:19,radford:19,raffel:20,brown:20}, Chen \textit{et. al.} \cite{chen:20} propose a knowledge-grounded pretraining framework (KGPT) trained on 1.8 million graph-text pairs of their knowledge-grounded dataset KGTEXT built with Wikipedia hyperlinks matched to WikiData \cite{vrandevcic:14}. The framework consists of a graph attention network \cite{velivckovic:18} based encoder and transformer \cite{vaswani:17} based encoders and decoders. Ke \textit{et. al.} \cite{ke:21} propose three graph-specific pretraining strategies based on the KGTEXT dataset - reconstructing masked narratives based on the input graph, conversely, reconstructing masked graph entities based on the narrative, and matching the graph and narrative embeddings with optimal transport. Similarly, Agarwal \textit{et. al.} \cite{agarwal:21} verbalize the entirety of the Wikidata Corpus \cite{vrandevcic:14} with two-step fine-tuning for T5 \cite{raffel:20} to construct their KeLM corpus. The authors utilize this corpus to train knowledge-enhanced language models for downstream NLG taks with significant improvements shown in the performance of REALM \cite{guu:20} and LAMA \cite{petroni:19} for both retrieval as well as question answering tasks. Similarly, specifically geared for logical inference from tables, Liu \textit{et. al.} \cite{plog:2022} propose PLoG wherein a PLM is first pre-trained on table-to-logic conversion intended to aid logical table-to-text generation for downstream datasets to the likes of LogicNLG.

\subsubsection{\underline{Autoencoders}}
For a D2T framework solely based on unlabeled text, Freitag and Roy \cite{freitag:18} adapt the training procedure of a denoising auto-encoder (DAE) \cite{vincent:08} to the seq2seq framework with the notion of reconstructing each training example from a partially destroyed input. For each training instance $x_{i}$, a percentage $p$ (sampled from a Gaussian distribution) of words are removed at random to get a partially destroyed version $\hat{x}_{i}$. However, the authors note carry over of this unsupervised approach across further D2T tasks such as the WebNLG challenge \cite{gardent:17} could be limited by the fact that the slot names in WebNLG contributes to the meaning representation.

\subsection{Innovations outside of the Seq2Seq Framework}
\subsubsection{\underline{Template Induction}} 
For enhanced interpretability and control in D2T, Wiseman \textit{et. al.} \cite{wiseman:18} propose a neural parameterization of the hidden semi-markov model (HSMM) \cite{murphy:02} that jointly learns latent templates with generation. With discrete latent states $z_{t}$, length variable $l_{t}$, and a deterministic binary variable $f_{t}$ that indicates whether a segment ends at time $t$, the HSMM is modeled as a joint-likelihood (\ref{shmm}). Further, associated phrases from $x$ can be mapped to latent states $z_{t}$ such that common templates can be extracted from the training dataset as a sequence of latent states $z^{i} = \{z_{1}^{i},...,z_{S}^{i}\}$. Thus, the model can then be conditioned on $z^{i}$ to generate text set to the template. Following this, Fu \textit{et. al.} \cite{fu:20} propose template induction by combining the expressive capacity of probabilistic models \cite{murphy:12} with graphical models in an end-to-end fashion using a conditional random field (CRF) model with Gumbel-softmax used to relax the categorical sampling process \cite{jang:17}. The authors note performance gains on HSMM-based baselines while also noting that neural seq2seq fare better than both.
\vspace{-0.2cm}
\begin{multline}
\label{shmm}
    P(y,z,l,f|x) = \prod_{t=0}^{T-1} P(z_{t+1},l_{t+1}|z_{t},l_{t},x)^{f_{t}}
    \times \prod_{t=1}^{T} P(y_{t-l_{t}+1:t}|z_{t},l_{t},x)^{f_{t}}
\end{multline}

\subsubsection{\underline{Discrete Neural Pipelines}}
While the traditional D2T pipeline observes discrete modeling of the content planning and linguistic realization stages \cite{reiterdale:97}, neural methods consolidate these discrete steps into end-to-end learning. With their neural referring expressions generator NeuralREG \cite{ferreira:18} appended to discrete neural pipelines, and the GRU \cite{cho:14} and transformer \cite{vaswani:17} as base models for both, Ferreira \textit{et. al.} \cite{ferreira:19} compare neural implementations of these discrete pipelines to end-to-end learning. In their findings, authors note that the neural pipeline methods generalize better to unseen domains than end-to-end methods, thus alleviating hallucination tendencies. This is also corroborated by findings from  Elder \textit{et. al.} \cite{elder:19}.

\subsubsection{\underline{Computational Pragmatics}}
Pragmatic approaches to linguistics naturally correct under-informativeness problems \cite{grice:75,horn:84} and are often employed in grounded language learning \cite{mao:16,monroe:17}. Shen \textit{et. al.} \cite{shen:19} adopt the reconstructor-based \cite{fried:18} and distractor-based \cite{cohn:18} models of pragmatics to MR-to-text generation. These models extend the \textit{base speaker} models $S_{0}$ using reconstructor $R$ and distractor $D$ based \textit{listener} models $L(y|x) \in \{L^{R},L^{D}\}$ to derive \textit{pragmatic speakers} $S_{1}(y|x) \in \{S_{1}^{R},S_{1}^{D}\}$ (\ref{prag:1},\ref{prag:2}) where $\lambda$ and $\alpha$ are rationality parameters controlling how much the model optimizes for discriminative outputs.
\begin{align}
    \label{prag:1}
    S_{1}^{R}(y|x) &= L^{R}(x|y)^{\lambda} \cdot S_{0}(y|x)^{1-\lambda} \\
    \label{prag:2}
    S_{1}^{D}(y|x) &\propto L^{D}(x|y)^{\alpha} \cdot S_{0}(y|x)
\end{align}

\section{Evaluation of Data-to-text Systems}
In this section, we take a deeper look into the specifics of evaluation for D2T systems. Traditionally, the evaluation of D2T systems is compartmentalized into either \textit{intrinsic} or \textit{extrinsic} measures \cite{belz:06}. The former either uses automated metrics to compare the generated narrative to a reference text or employs for human judgement \cite{joshi:15} - both evaluating the properties of the system output. The latter focuses on the ability of the D2T system to fulfill its intended purpose of imparting information - to what degree does the system achieve its overarching task for which it was developed. From the analysis of 79 papers spanning 2005-2014, Gkatzia and Mahamood \cite{gkatzia:15} note the overwhelming prevalence of intrinsic evaluation with 75.7\% of articles reporting it compared to 15.1\% that report an extrinsic measure. This is unsurprising, as intrinsic evaluation can be automated and is often convenient, not requiring additional crowd-sourced human labor and collection of feedback from deployed systems. As such, Reiter \cite{reiter:blog} notes the importance of extrinsic (pragmatic) evaluation and its absence in the field. Thus, the absence of literature in extrinsic evaluation measures leads us to focus on the innovations in improving the quality of intrinsic evaluation metrics (\textsection 6.2). For a broader view on the evaluation of text generation systems in the greater NLG landscape, we refer the readers to a recent survey of evaluation practices for text generation systems by Celikyilmaz et. al. \cite{celikyilmaz:20}.

\subsection{BLEU: the false prophet for D2T}
With the abundance of paired datasets where each data instance is accompanied by a human generated reference text, often referred to as the \textit{gold standard}, the NLG community has sought after quick, cheap, and effective metrics for evaluation of D2T systems. The adoption of automated metrics such as BLEU, NIST, and ROUGE, by the machine translation (MT) community, by the virtue of their correlation with human judgement \cite{papineni:02, doddington:02, lin:03}, similarly carried over to the D2T community. Among them, Belz and Gatt \cite{belz2:08} note that NIST best correlates with human judgements on D2T texts, when compared to 9 human domain-experts and 21 non-experts. However, they note that these n-grams based metrics perform poorer in D2T as compared to MT due to the domain-specific nature of D2T systems wherein the generated texts are judged better by humans than human-written texts.

From a review of 284 correlations reported in 34 papers, Reiter \cite{reiter:18} notes that the correlations between BLEU and human evaluations are inconsistent - even in similar tasks. While automated metrics can aid in the diagnostic evaluation of MT systems, the author showcases the weakness of BLEU in the evaluation of D2T systems. This notion has been resonated several times \cite{scott:07, reiter:09}. On top of this, undisclosed parameterization of these metrics and the variability in the tokenization and normalization schemes applied to the references can alter this score by up to 1.8 BLEU points for the same framework \cite{post:18}. Similarly, it has also been shown that ROUGE tends to favor systems that produce longer summaries \cite{sun:19}. Further complicating the evaluation of D2T is the fact modern frameworks are neural - comparing score distributions, even with the aid of statistical significance tests, are not as meaningful due to the non-deterministic nature of neural approaches and accompanying randomized training procedures \cite{reimers:17}.

\subsection{Innovations in Intrinsic Evaluation}
Noting the shortcomings of prevalent word-overlap metrics (\textsection 6.1), alternative automated metrics for intrinsic evaluation have been proposed (\textsection 6.2.1 \& \textsection 6.2.2). To account for divergence in reference texts, Dhingra \textit{et. al.} \cite{dhingra:19} propose PARENT - a metric that computes precision and recall of the generated narrative $\hat{y}$ with both the gold narrative $y$ and its entailment to the semi-structured tabular input $x$.

\subsubsection{\underline{Extractive Metrics}}
With dialogue generation models adopting classification-backed automated metrics \cite{kannan:16, li:17}, Wiseman \textit{et.  al.} \cite{wiseman:17} propose a relation extraction system to the likes of \cite{collobert:11,dos:15} wherein the record type $r.t$ is predicted using its corresponding entity $r.e$ and value $r.m$ as $p(r.t | e,m;\theta)$. With such a relation extraction system, the authors propose three metrics for automated evaluation:
\begin{itemize}
    \item \textit{Content Selection} (CS) is represented by the precision and recall of unique relations extracted from $\hat{y}_{1:t}$ that are also extracted from $y_{1:t}$.
    \item \textit{Relation Generation} (RG) is represented by the precision and number of unique relations extracted from $\hat{y}_{1:t}$ that can be traced to $x$.
    \item \textit{Content Ordering} (CO), similarly, by the normalized Damerau-Levenshtein distance \cite{brill:00} between the sequence of records extracted from $y_{1:t}$ and $\hat{y}_{1:t}$.
\end{itemize}
The authors note that, given the two facets of D2T, CS pertains to \textit{what to say} and CO to \textit{how to say it} while RG pertains to both (factual correctness).

\subsubsection{\underline{Contextualized Metrics}}
B{\"o}hm \textit{et. al.} \cite{bohm:19} note that while modern frameworks for text generation compete with higher scores on automated word-overlap metrics, the quality of the generation leaves a lot to be desired. As such, the adaptation of continuous representations based metrics shifts the focus from surface-form matching to semantic matching. Zhang \textit{et. al.} \cite{zhang:19} introduce BERTScore which computes similarity scores for tokens in the system and reference text based on their BERT \cite{devlin:19} embeddings while Mathur \textit{et. al.} \cite{mathur:19} devise supervised and unsupervised metrics for NMT based on the same BERT embeddings - both having substantially higher correlation to human judgement compared to standard word-overlap metrics (see \textsection 6.1). Following this, Clark \textit{et. al.} \cite{clark:19} extend the word mover's distance \cite{kusner:15} to multi-sentence evaluation using ELMo representations \cite{peters:18}. Zhao \textit{et. al.} \cite{zhao:19} propose the MoverScore which uses contextualized embeddings from BERT where the aggregated representations are computed based on power means \cite{ruckle:18}. Du{\v{s}}ek and Kasner \cite{duvsek:20} employ RoBERTa \cite{liu:19} for natural language inference (NLI), where, for a given \textit{hypothesis} and \textit{premise}, the model computes scores for \textit{entailment} between the two. While lower scores for forward entailment can point to omissions, backward entailment scores correspondingly point to hallucinations. Similarly, Chen \textit{et. al.} \cite{chen_c:20} propose parsing-based and adversarial metrics to the evaluate model correctness in logical reasoning.

\subsubsection{\underline{Human Judgement}}
While human judgment is often considered to be the ultimate D2T evaluation measure, they are subject to a high degree of inconsistency (even with the same utterance), which may be attributed to the judge's individual preferences \cite{walker:07,dethlefs:14} - an issue that could be circumvented through a larger sample size, however such an endeavor is accompained with an equal increase in cost for data acquisition. As such, there have been several recommendations on the proper usage of ratings and Likert scales \cite{knapp:90, joshi:15, johnson:16}. From the analysis of 135 papers from specialized NLG conferences, Amidei \textit{et. al.} \cite{amidei:19} note that several studies employ the Likert scale on an item-by-item basis in contrast to its design as an aggregate scale, and the analysis performed on these scales with parametric statistics do not disclose the assumptions about the distribution of the population probability. Howcroft \textit{et. al.} \cite{howcroft:20}, on the other hand, note that the definitions of \textit{fluency} and \textit{accuracy} for which these scales are employed lack consistency among the papers investigated. To mitigate these inconsistencies through good experimental design, Novikova \textit{et. al.} \cite{novikova:18} propose RankME, a relative-ranking based magnitude estimation method that combines the use of continuous scales \cite{belz:11, graham:13}, magnitude estimation \cite{siddharthan:12}, and relative assessment \cite{callison:07}. Further, to address the quadratic growth of data required for cross-sytem comparisons, the authors adopt TrueSkill \cite{herbrich:06}, a bayesian data-efficient ranking algorithm used in MT evaluation \cite{bojar:16}, to RankME. The HumEval workshop \cite{humeval_1, humeval_2, humeval_3} has been an invaluable resource in investigating the shortcomings and building maps to better practices in human evaluations.

\subsection{Emphasis on Reproducibility}

The last half-decade has seen the ML community place significant emphasis on the \textit{reproducibility} of academic results \cite{ml_repro, ml_repro_2}. However, the focus of these reproducibility efforts are placed on automated metrics (\textsection 6.1, \textsection 6.2.1, \textsection 6.2.2) with the reproducibility of human evaluation results receiving far less attention. As human evaluation is often considered the ultimate measure of D2T, Belz \textit{et. la.} \cite{reprogen} initiate \textit{ReproGen}, a shared task focused on reproducing the results of human evaluations - intended to shed better light on the reproducibility of human evaluations and the possible interventions in design and execution of human evaluations to make them more reproducible. The authors note the inconsistency (expected) of the evaluators across different studies and hence point towards the use of metadata standardization through data-sheets such as HEDS \cite{heds}. Providing a complementary view, van der Lee \textit{et. al.} \cite{best_practices} review practices in human evaluation from 304 publications in the International Conference in Natural Language Generation (INLG) and the Annual Meeting of the Association of Computation Linguistics (ACL) from 2018 to 2019 and outline the severe discrepancy in the spectrum of evaluator demographics and sample sizes, design practices, evaluation criteria and put forth some common ground through a set of best practices for conducting human evaluations.

\section{Conclusion and Future Directions}
As delineated in \textsection 3 \& \textsection 4, innovations in D2T take inspiration from several facets of NLG and ML. From alterations to the seq2seq, pretraining-finetuning, autoencoding, ensemble-learning, and reinforcement-learning paradigms, to domain-specific data preprocessing and data encoding strategies, the prospects for innovations in D2T appear as grand as that for the NLG landscape itself. Alongside, progresses made in non-anglocentric datasets, datacards that reinforce accountability, and metrics that offer heuristic evaluation, aid in elevating D2T standards. As NLG research evolves, so will D2T, and vice-versa. In the following sections, we impart our thoughts for future directions for each facet of D2T - the desiderata for D2T dataset design (\textsection 7.1), a forward look at the possibilities for approaches and architectures for D2T (\textsection 7.2) and finally, closing thoughts on the future of D2T evaluation (\textsection 7.3).

\subsection{Desiderata for D2T Datasets}
 In \textsection 2, we outlined the development of parallel corpora with data-narrative pairs alongside dominant benchmark datasets in each task category. In addition to these benchmarks, it is as crucial to acknowledge niche datasets - Obeid and Hoque \cite{obeid:20} compile a collection of 8,305 charts with their respective narratives, followed by Chart-to-text \cite{kanthara:22}, that encompasses 44,096 multi-domain charts. The shared gains and pitfalls in dataset design across the D2T task categories, as discussed in \textsection 2, offer insights that can aid the construction of future datasets with the potential to challenge the current paradigm: 
\begin{itemize}
    \item \textbf{Domain Agnosticism}: Although domain-specific datasets allow the models to learn and leverage domain-specific conventions for performance gains in niche tasks, the resulting models are less malleable to unseen domains. To be adaptable and deployable for unseen niche tasks that may vary based on user requirements, it is crucial that the datasets used to train D2T models are not restricted to a single domain to avoid over-fitting on domain-specific keywords.
    \item \textbf{Dataset Consistency}: Often, the greatest challenges for D2T systems, namely \textit{hallucination} and \textit{omission} (see \textsection 3.2), can be traced back to the datasets. Datasets facing divergence (as outlined in \textsection 2.3), wherein the narratives are not consistent with the data instances or vice-versa, often lead to models that hallucinate or omit important aspects of the data \cite{rohrbach:18}.
    \item \textbf{Human-crafted References}: Often, to replicate human linguistics, datasets in NLP/NLG contain human annotations (narratives), considered as \textit{gold} references. Reiter \cite{reiter:blog} notes that D2T datasets, unintentionally, may contain machine-generated annotations, such as those for WeatherGov, and urges the community to focus on human-centered narratives.
    \item \textbf{Linguistic Diversity}: It is vital to acknowledge that the majority of the D2T benchmark datasets are anglo-centric. Joshi \textit{et. al.} \cite{joshi:20} note that models built on non-anglo-centric datasets, which are fewer and far between, have the potential to impact many more people than models built on highly resourced languages. The WebNLG 2020 challenge\footnote{\url{https://webnlg-challenge.loria.fr/challenge_2020/}}, for instance, encourages submissions for both English and Russian parsing.
\end{itemize}

\subsection{Approaches to Data-to-text Generation: Looking Forward}
 In the above sections \textsection 4 and \textsection 5, we have extensively outlined the recent innovations in D2T both inside and outside of seq2seq modeling. However, looking forward, with the emergence of highly capable large language models (LLMs) such as ChatGPT \cite{chatgpt}, below we discuss the reconciliation of these emergent technologies with the current D2T paradigm:
\begin{itemize}
    \item \textbf{Few-shot Learning}: Data-to-text generation is a task that requires extrapolation beyond general linguistic understanding and commonsense reasoning, thus general LLM prompting strategies \cite{cot} may not be suited for this endeavor. Adding to that the data sparsity prevalent in D2T, extensions of prefix-tuning \cite{li_b:21} and in-context sample search \cite{liu:21} may be especially favorable for building strong subset of samples for few-shot learning in LLMs.
    \item \textbf{Deviation from Task-specific Architectures:} The paradigm for data-to-text generation, as it stands now, prefers custom architectures, and rightfully so - they allow focused modeling of entities (\textsection 5.1.1) and dedicated attention mechanisms (\textsection 5.1.2 \& \textsection 5.1.6) to combat data infidelity that occurs as a consequence of RNN's lack of long-form coherence. Transformer-based LLMs, however, may inadvertently model these dependencies and attention mechanisms as a function of their self-attention modules, thus allowing the convergence towards a universal architecture.
    \item \textbf{Effective Linearization:} In line with the point above, the preference for plan-based approaches to D2T (\textsection 5.1.3) stems from issues in coherence. While there has been extensive work in linearizing graphs and tables (\textsection 4.2) showcasing that linearization in LLMs can be as effective, if not more, compared to dedicated encoders designed to capture inductive biases (\textsection 5.1.1, \textsection 5.1.2, \textsection 5.1.5), recent work suggests that LLMs can handle unexpected tasks simply through their linear transcription into linguistic sentences (the LIFT framework) \cite{lift}. This line of work has extensive potential for modeling plans through their transcriptions into sentences.
    \item \textbf{Numeracy for Data-to-text Generation:} The NLP niche of building LLMs capable of quantitative reasoning (often referred to as Math-AI or Math-NLP) has garnered significant interest from the research community \cite{num:1,num:2,num:3}. Although there are works in D2T that incorporate this aspect \cite{gong_b:20}, the two research niches are often disparate. D2T, a field that aims to combat hallucination of data points, has a lot to gain from the advances in Math-NLP that enable models to better reason about said data points.
    \item \textbf{Interfacing with External APIs:} In line with the above point, interfacing LLMs with computational APIs (Wolfram Aplha \cite{wolfram:2023}, ToolFormer \cite{toolformer:2023}) has showcased significant enhancement potential for these already capable models. This paves a path in D2T where we deviate from viewing the generation of narratives as a sequential input-to-output mapping but rather a more involved loop comprising of numerical and logical reasoners, computational engines, pattern matchers, and validators that combine to form the greater NLG pipeline.
\end{itemize}

\subsection{The Future of D2T Evaluation}
While NLP systems generally have benchmark datasets that closely resemble the target tasks that these systems are intended to be deployed for (summarization, sentiment analysis, and language translation), data-to-text systems are highly specialized to the incoming data stream which differs from user to user. Thus, a one-size-fits-all approach to benchmarking, especially with automated metrics on bechmark datasets, can not showcase the utility of these systems in the real world, leading to an urgency for practical tools for extrinsic evaluation. Additionally, besides the need for fluidity and fidelity, systems placed in the real world require accountability \cite{mitchell:19}. The current leaderboard system poses the risk of blind metric optimization with disregard to model size and fairness \cite{ethayarajh:20}. For a holistic approach to evaluation, Gehrmann \textit{et. al.} \cite{gehrmann:21} propose a living benchmark, GEM, similar to likes to Dynabench \cite{kiela:21}, providing challenge sets (a set of curated test sets intended to be challenging) and benchmark datasets accompained with their D2T-specific data cards \cite{bender:18}.

Further, as unified D2T frameworks become more decentralized with growing user-bases, the designers of these systems can utilize the user-interaction logs as measures for extrinsic evaluation, similar to the likes of ChatGPT \cite{chatgpt}. While the D2T community places greater emphasis on measures to evaluate the \textit{quality} of the generated narrative, the \textit{utility} of these narratives can be evaluated with task effectiveness with and without the presence of narratives \cite{causality:1, causality:2}.  

\section{Acknowledgements}
This work is supported in part by US National Science Foundation grant DGE1545362.

\bibliographystyle{ACM-Reference-Format}
\bibliography{custom}

\end{document}